\documentclass[10pt,twocolumn,letterpaper]{article}

\usepackage{cvpr}              
\usepackage[ruled,vlined]{algorithm2e}
\usepackage{xcolor}
\usepackage{mathtools}

\definecolor{cvprblue}{rgb}{0.21,0.49,0.74}
\usepackage[pagebackref,breaklinks,colorlinks,allcolors=cvprblue]{hyperref}
\usepackage{multirow}
\usepackage{amssymb}
\usepackage{capt-of}

\def\paperID{2118} 
\def\confName{CVPR}
\def\confYear{2026}

\title{PhysGen: Physically Grounded 3D Shape Generation for Industrial Design}

\author{
Yingxuan You \quad
Chen Zhao \quad
Hantao Zhang \quad
Ming Xu \quad
Pascal Fua \\
CVLab, EPFL \\
{\tt\small \{yingxuan.you, chen.zhao, hantao.zhang, mingda.xu, pascal.fua\}@epfl.ch}
}

\begin{document}
\twocolumn[
\begin{@twocolumnfalse}
\maketitle
\begin{center}
    \centering
    \vspace{-4mm}
    \includegraphics[width=\linewidth]{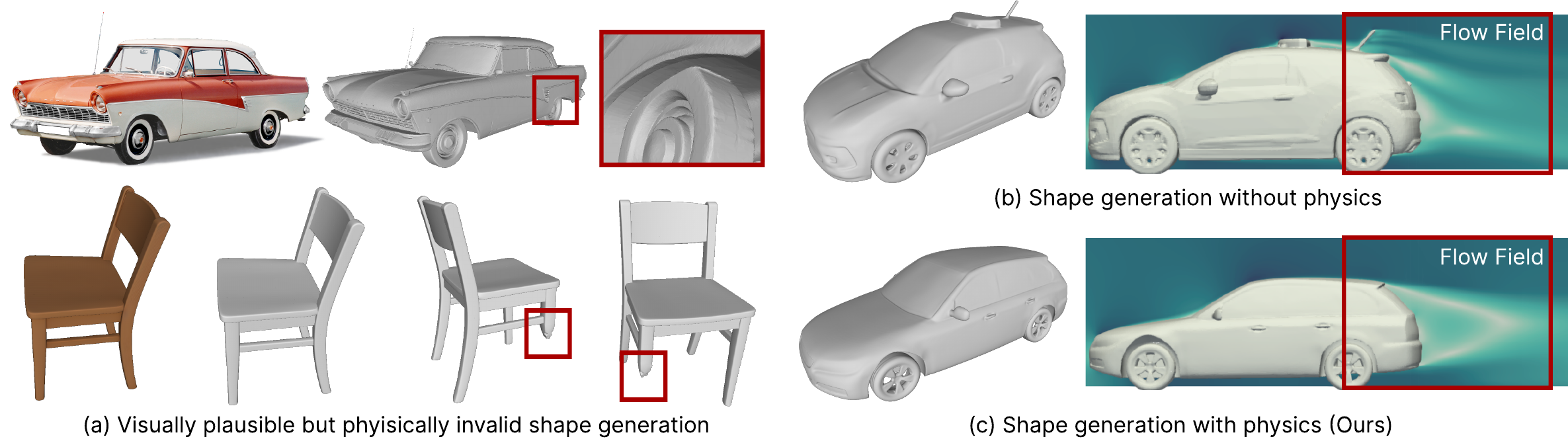}
    \vspace{-6mm}
    \captionof{figure}{Physical knowledge is crucial for realistic and functionally efficient 3D shape generation. Without physics, (a) generated shapes may appear visually plausible yet violate basic physical feasibility, such as car wheels intersecting the body or chairs with broken or unstable legs, and (b) aerodynamic shapes produce wide turbulent wakes, indicating low aerodynamic efficiency. (c) In contrast, physics-guided generation produces shapes with coherent flow and reduced drag, achieving both aesthetic appeal and physical efficiency.}
    \label{fig:fig1}
    \vspace{2mm}
\end{center}
\end{@twocolumnfalse}
]

\newif\ifdraft
\drafttrue

\definecolor{burntorange}{rgb}{0.8, 0.33, 0.0}
\definecolor{orange}{rgb}{1,0.5,0}
\definecolor{green0}{rgb}{0.1,0.7,0.1}
\definecolor{blue}{rgb}{0,0.5,1}

\ifdraft
\newcommand{\PF}[1]{{\color{red}{\bf PF: #1}}}
\newcommand{\pf}[1]{{\color{red} #1}}
\newcommand{\YXY}[1]{{\color{green0}{\bf YX: #1}}}
\newcommand{\yxy}[1]{{\color{green0} #1}}
\newcommand{\CZ}[1]{\textcolor{SpringGreen}{\textbf{CZ: #1}}}
\newcommand{\cz}[1]{{\textcolor{SpringGreen} #1}}
\newcommand{\MX}[1]{{\color{orange}{\bf MX: #1}}}
\newcommand{\mx}[1]{{\color{orange} #1}}
\newcommand{\HTZ}[1]{{\color{blue}{\bf HT: #1}}}
\newcommand{\htz}[1]{{\color{blue} #1}}
\newcommand{\todo}[1]{{\color{red}#1}}
\newcommand{\TODO}[1]{\textbf{\color{red}[TODO: #1]}}

\else
\newcommand{\PF}[1]{{\color{red}{}}}
\newcommand{\pf}[1]{ #1 }
\newcommand{\YXY}[1]{{\color{green0}{}}}
\newcommand{\yxy}[1]{ #1 }
\newcommand{\CZ}[1]{\textcolor{SpringGreen}{}}
\newcommand{\cz}[1]{{ #1}}
\newcommand{\MX}[1]{{\color{orange}{}}}
\newcommand{\mx}[1]{{ #1}}
\newcommand{\HTZ}[1]{{\color{blue}{}}}
\newcommand{\htz}[1]{{#1}}
\newcommand{\TODO}[1]{}
\newcommand{\todo}[1]{#1}
\fi

\newcommand{\acron}[0]{{\it PhysRec}}

\newcommand{\parag}[1]{\vspace{-2mm}\paragraph{#1}}
\newcommand{\sparag}[1]{\vspace{-2mm}\subparagraph{#1}}
\newcommand{\subsec}[1]{\vspace{-0.4mm}\subsection{#1}}

\newcommand{\fdyn}[0]{{f_{\rm dyn}}}
\newcommand{\gdyn}[0]{{g_{\rm dyn}}}
\newcommand{\gcnn}[0]{{\tilde{g}_{\rm dyn}}}

\newcommand{\mL}[0]{\mathcal{L}}
\newcommand{\mM}[0]{\mathcal{M}}
\newcommand{\mS}[0]{\mathcal{S}}
\newcommand{\mZ}[0]{\mathcal{Z}}
\newcommand{\mD}[0]{\mathcal{D}}
\newcommand{\mX}[0]{\mathcal{X}}

\newcommand{\sR}[0]{\mathbb{R}}

\newcommand{\bl}[0]{\mathbf{\Lambda}}
\newcommand{\bt}[0]{\mathbf{\Theta}}

\newcommand{\bc}[0]{\mathbf{c}}
\newcommand{\bd}[0]{\mathbf{d}}
\newcommand{\bC}[0]{\mathbf{C}}
\newcommand{\bM}[0]{\mathbf{M}}
\newcommand{\bQ}[0]{\mathbf{Q}}
\newcommand{\bR}[0]{\mathbf{R}}
\newcommand{\bu}[0]{\mathbf{u}}
\newcommand{\bU}[0]{\mathbf{U}}
\newcommand{\bo}[0]{\mathbf{o}}
\newcommand{\bp}[0]{\mathbf{p}}
\newcommand{\br}[0]{\mathbf{r}}
\newcommand{\bs}[0]{\mathbf{s}}
\newcommand{\bv}[0]{\mathbf{v}}

\newcommand{\bN}[0]{\mathbf{N}}
\newcommand{\bD}[0]{\mathbf{D}}
\newcommand{\bS}[0]{\mathbf{S}}

\newcommand{\bX}[0]{\mathbf{X}}
\newcommand{\bx}[0]{\mathbf{x}}
\newcommand{\by}[0]{\mathbf{y}}
\newcommand{\bY}[0]{\mathbf{Y}}
\newcommand{\bw}[0]{\mathbf{w}}
\newcommand{\bW}[0]{\mathbf{W}}
\newcommand{\bz}[0]{\mathbf{z}}
\newcommand{\bZ}[0]{\mathbf{Z}}

\newcommand{\ours}[0]{{\it AlterDesign}}

\begin{abstract}
Existing generative models for 3D shapes can synthesize high-fidelity and visually plausible shapes. For certain classes of shapes that have undergone an engineering design process, the realism of the shape is tightly coupled with the underlying physical properties, e.g., aerodynamic efficiency for automobiles. Since existing methods lack knowledge of such physics, they are unable to use this knowledge to enhance the realism of shape generation. Motivated by this, we propose a unified physics-based 3D shape generation pipeline, with a focus on industrial design applications. Specifically, we introduce a new flow matching model with explicit physical guidance, consisting of an alternating update process. We iteratively perform a velocity-based update and a physics-based refinement, progressively adjusting the latent code to align with the desired 3D shapes and physical properties. We further strengthen physical validity by incorporating a physics-aware regularization term into the velocity-based update step. To support such physics-guided updates, we build a shape-and-physics variational autoencoder (SP-VAE) that jointly encodes shape and physics information into a unified latent space. The experiments on three benchmarks show that this synergistic formulation improves shape realism beyond mere visual plausibility. Our code and model weights are available at \href{https://github.com/kasvii/PhysGen}{\texttt{https://github.com/kasvii/PhysGen}}.
\end{abstract}
    

\section{Introduction}
\label{sec:intro}
Generative AI has achieved remarkable success in text~\cite{minaee2024large} and image~\cite{Yazdani25a} processing. Meanwhile, 3D content creation has gained increasing attention with applications across a wide range of fields, including virtual reality \cite{Ma25a}, gaming~\cite{Chen25d}, retail~\cite{Chen25e}, and engineering design~\cite{Vatani25a}. The existing 3D generative models~\cite{Zhang23d, Chen25b, Chan22b, Li25b} excel in creating apparently realistic objects. However, such realism is only skin-deep. Close examination often reveals inconsistencies or implausibilities, which might be acceptable in some entertainment applications where the objective is merely to produce visually appealing results, but become detrimental when the end goal is engineering-oriented. For example, as illustrated in Fig.~\ref{fig:fig1}, the wheels of generated cars~\cite{Li25b} touch the body of the car, which makes no sense when designing functional cars. Similarly, the chairs~\cite{Lai25a} generated from images contain feet with the wrong topology, making them unsuitable to bear the weight of a person.  

These failings can be traced to the fact that these methods are trained solely on static datasets~\cite{Deitke23a, Chang15} of 3D shapes, without regard to the engineering design process that was used to create them. For instance, automobile and airplane designs~\cite{Regenwetter22a, Song24a} are usually optimized for aerodynamic efficiency, which conditions their shape beyond just aesthetics. Such physical awareness is absent from the current 3D shape generation pipelines~\cite{Chan22b, Li25b, Lai25a}. To bridge this gap, we demonstrate that physical knowledge can be used to define regularization constraints, which are then used to improve the quality and realism of 3D shape generation.

To this end, we present \emph{PhysGen}, an approach to enforcing physical realism on generated 3D shapes, by using information around the physics of the shape. 
Specifically, we propose a physics-guided flow matching model, built upon an alternating update algorithm that switches between a velocity-based update and a physics-based refinement.
Given the desired physical properties, we further enhance the physical guidance by introducing a physics-aware regularization in the velocity-based update. We update the latent code leveraging gradients derived from the discrepancy between the predicted and target physical values. Notably, 3D generative models typically rely on VAE-learned latent spaces, but existing shape VAEs~\cite{Zhang23d, Chen25b} encode no physics, making physical properties unrecoverable from their latent codes. To enable effective physics-based regularization, we propose a shape-and-physics variational autoencoder (SP-VAE) that embeds both 3D shape and physics information into a unified latent space.

In this paper, we focus on the application of 3D shape generation in automobile design, where physical performance is pivotal to ensuring that generated shapes satisfy real-world engineering requirements. We also showcase additional applications, such as structural optimization~\cite{Zhan25a} under prescribed loads and boundary conditions, which demonstrates the generalization ability of our approach. We evaluate our \emph{PhysGen} across benchmarks~\cite{Elrefaie24a, Chang15}, including unconditional 3D generation, 3D generation conditioned on a sketch, and 3D generation conditioned on a real single-view image. The results demonstrate that by utilizing physics-based regularization, \emph{PhysGen} produces shapes that are more geometrically plausible and physically efficient than previous approaches. We also conduct comprehensive ablation studies on the key components in our pipeline. 
In summary, our contributions are as follows:

\begin{itemize}
    \item We investigate the physical realism in 3D shape generation models and propose a novel physics-guided flow matching model that generates physically efficient and aesthetically pleasing 3D shapes. 

    \item We achieve the physical guidance by alternating between a velocity-based update with physics-aware regularization and a physics-based refinement.
    
    \item We build a shape-and-physics variational autoencoder that encodes 3D shape and physics in a unified latent space, enabling the physical guidance in the presented flow matching model.
\end{itemize}

\section{Related Work}
\label{sec:relatedwork}

\subsection{3D Shape Generation}
Inspired by advances in image and video generation, generative models for 3D shape generation~\cite{Zhang23d, Chen25b, Li25b, Wu24b, Zhang24a, Zhao23Michelangelo} have recently made rapid progress. 
One line of work distills knowledge from powerful 2D diffusion models into 3D domains~\cite{Chen23a, Li23h, Lin23a, Liu24a, Poole00a, Qiu24a, Tang00a}, alleviating the scarcity of high-quality 3D data. While effective at producing visually appealing shapes, these methods often exhibit slow convergence and unstable optimization.
Another line of work builds native 3D generative models~\cite{Zhang23d, Chen25b, Li25b, Zhang24a}. Methods such as 3DShape2VecSet~\cite{Zhang23d} and Dora~\cite{Chen25b} commonly adopt a two-stage paradigm, where a 3D VAE first encodes geometry into a latent space, followed by a diffusion model for 3D shape generation.
While achieving visually plausible shapes, neither the VAE nor the diffusion models encodes physics information, which limits their usefulness in physics-sensitive domains. To address this, we propose a joint shape-physics VAE that embeds geometry and physics information into a unified latent space, capturing their intrinsic correlations and enabling more physically aware shape generation.

\subsection{Physics-Aware Shape Generation}
Physics is often incorporated into 3D generation via post-processing optimization~\cite{Pastrana23a, Guo24d}. Finite element methods (FEMs)~\cite{Lee25a, Xue23a, Wu23b} and force density methods (FDMs)~\cite{Pastrana23a} can stabilize generated meshes under external forces~\cite{Guo24d}, but they operate solely on explicit meshes. In contrast, generative models encode shape priors implicitly in latent spaces~\cite{Zhang23d, Chen25b, Li25b}, making optimization without shape priors highly sensitive to initialization and prone to failure once shapes drift out of distribution~\cite{Guo24d}.
These limitations motivate physics-aware generation directly in latent space. TripOptimizer~\cite{Vatani25a} applies drag-driven gradient updates to VAE parameters, but often disrupts the model priors learned from the dataset. Recent works~\cite{Maze23a, Giannone23a} inject physical gradients into the diffusion process. However, physics estimates on noisy early samples are unreliable, and the limited steps in later diffusion stages are insufficient for convergence. To address these issues, we propose a physics-guided 3D shape generation method that integrates flow-matching generation and physical guidance into a unified framework. Rather than generating a shape followed by heavy post-optimization, our method alternates between a velocity-based update and a physical refinement, facilitating the convergence toward shapes that are both geometrically plausible and physically valid.
\begin{figure*}
    \centering
    \includegraphics[width=0.92\linewidth]{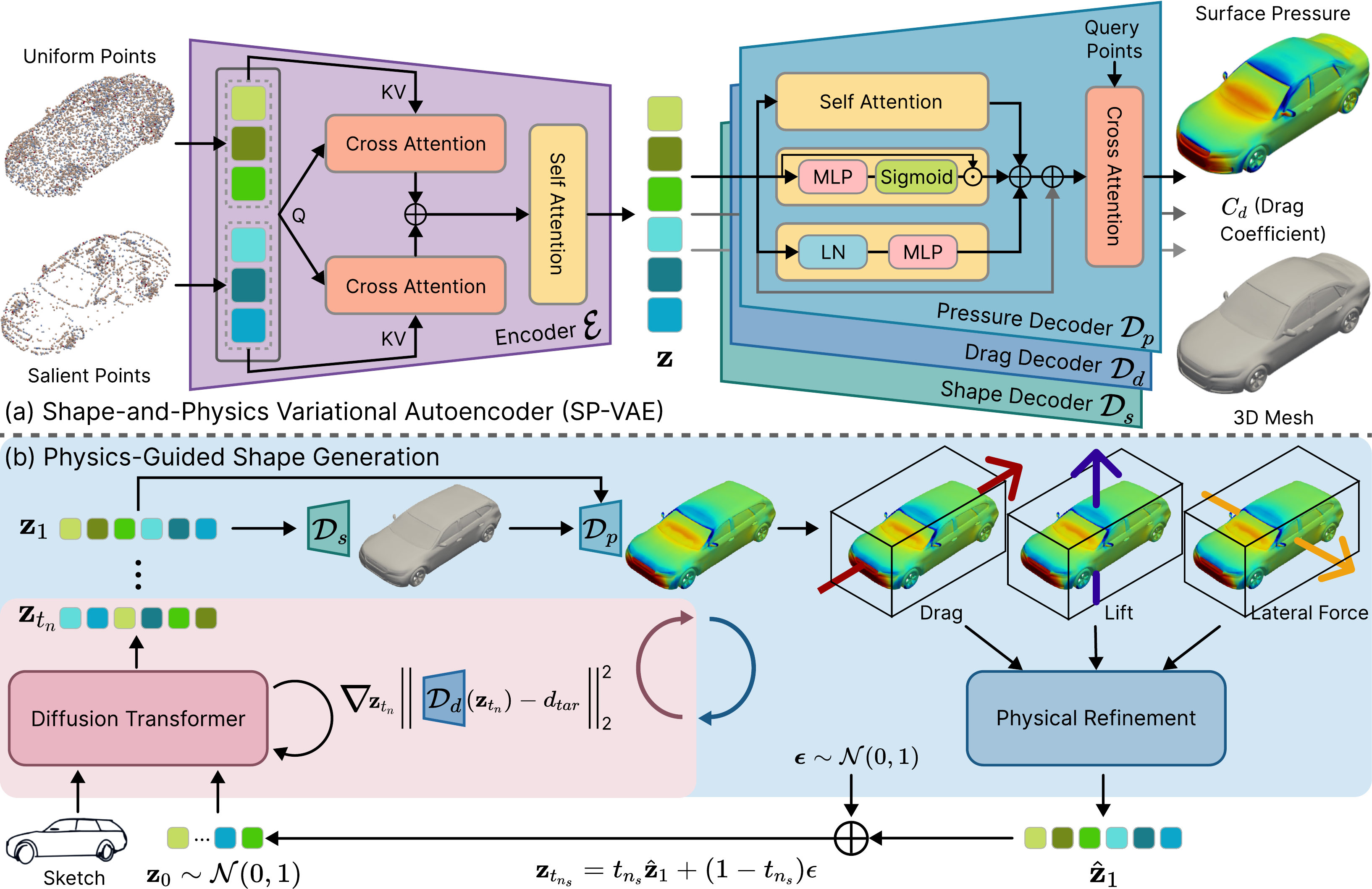}
    \vspace{-2mm}
    \caption{Overview of the proposed framework. (a) The proposed SP-VAE learns a unified latent representation that jointly encodes geometric structure and physical properties. From this shared representation, three decoders reconstruct the 3D shape, surface pressure field, and drag coefficient, respectively. (b) The physics-guided shape generation iteratively bridges flow-matching updates and physical refinements, optionally conditioned on an image, such as a sketch. This alternating strategy updates the latent code to align with the desired 3D shape and physical properties, ensuring both visual plausibility and physical validity.
    }
    \vspace{-2mm}
    \label{fig:overview}
\end{figure*}

\section{Method}
\label{sec:method}

\noindent \textbf{Problem Formulation.} We aim to generate a 3D shape $\mathcal{S}$, conditioned on physics information $A$ and, optionally, on a 2D image $\textbf{I}$. In other words, we aim to maximize the posterior distribution $q(\mathcal{S}| A)$ (or $q(\mathcal{S}| \textbf{I}, A)$ when $\textbf{I}$ is available). We use a flow matching model~\cite{Liu23e} to evolve the samples based on physical guidance to ensure that the resulting shapes are plausible from a physics point of view. In this paper, we take $A$ to be aerodynamic properties formulated as $A \coloneqq \{C_d, \mathcal{P}\}$, where $C_d$ is the dimensionless coefficient of drag and $\mathcal{P}:\mathbb{R}^3 \rightarrow \mathbb{R}^+$ is the pressure field measured at surface points. $\mathcal{S}$ is represented by a signed distance function (SDF) $S : \mathbb{R}^{3} \rightarrow \mathbb{R}$ that evaluates the signed distance of 3D points to the surface. 

\noindent \textbf{System Overview.} To achieve the mapping from physics information to a 3D shape, we propose a unified framework that incorporates physical awareness into the flow matching generation mechanism, as illustrated in Fig.~\ref{fig:overview}. Specifically, we develop a joint latent space capturing both geometry and physical properties (Sec.~\ref{sec:vae}), and a physics-guided flow matching model for 3D shape generation (Sec.~\ref{sec:opt}).

\subsection{Shape-and-Physics Variational Autoencoder}
\label{sec:vae}
Recall that we aim to achieve physics-guided 3D shape generation. However, the existing 3D shape VAEs~\cite{Chen25b, Zhang23d} only encode 3D geometry information in the latent space. The lack of physical properties in these representations makes it difficult to generate physically efficient 3D shapes from the latent space. To this end, we develop a Shape-and-Physics Variational Autoencoder (SP-VAE) that jointly encodes shape and physics within a unified latent space.

\subsubsection{Shape Encoder and Decoder}
We adopt Dora~\cite{Chen25b} as the baseline architecture for our shape encoder $\mathcal{E}$ and shape decoder $\mathcal{D}_s$. Specifically, uniform surface points $\mathbf{P}_u$ and salient edge points $\mathbf{P}_s$ are first extracted from the mesh as inputs.
The encoder applies a dual cross-attention mechanism between $\mathbf{P}_u$ and $\mathbf{P}_s$, aggregates the results via self-attention, and outputs the latent code $\mathbf{z}$.
For shape decoding, different from Dora that predicts an occupancy field, we adopt an SDF representation to capture finer geometric details~\cite{Li25c}. Several self-attention layers process $\mathbf{z}$, followed by cross-attention with a linear projection that takes query point $\mathbf{x} \in \mathbb{R}^{3}$ and outputs corresponding SDF value $s = \mathcal{D}_s(\mathbf{x}, \mathbf{z})$. The final 3D mesh is reconstructed via the marching cubes algorithm~\cite{Lorensen98a}.
Detailed architecture is provided in the supplementary material.

\subsubsection{Pressure and Drag Decoders}
\noindent \textbf{Pressure Decoder $\mathcal{D}_p$.} 
The pressure decoder predicts a continuous pressure field that allows querying the pressure value $p \in \mathbb{R}$ based on the latent code $\mathbf{z}$ at any 3D spatial point $\mathbf{x} \in \mathbb{R}^{3}$, formulated as:
\vspace{-2mm}
\begin{equation}
    p = \mathcal{D}_p(\mathbf{x}, \mathbf{z})
\vspace{-2mm}
\end{equation}
As shown in Fig.~\ref{fig:overview} (top right), given the latent code $\mathbf{z}$, the decoder employs three parallel branches to capture multi-level physical information: a self-attention models global surface dependencies, a channel branch uses a squeeze-excitation~\cite{Hu18Squeeze} mechanism to reweight feature channels, 
and an MLP refines local representations.
The outputs of these branches are fused as 
$
    \mathbf{z}_\text{fused} = w_1 \cdot \mathbf{z}_\text{attn} + w_2 \cdot \mathbf{z}_\text{channel} + w_3 \cdot \mathbf{z}_\text{mlp}
$,
where $w_1$, $w_2$ and $w_3$ are learnable weights.
Finally, the fused features are decoded through a cross-attention layer that takes 3D spatial point $\mathbf{x}$ as queries and predicts the corresponding pressure value $p$.

\noindent \textbf{Drag Coefficient Decoder $\mathcal{D}_d$.} 
Unlike the previous decoders that estimate spatial fields, the drag decoder predicts a global drag coefficient $C_d \in \mathbb{R}$ from each latent code. Similar to the pressure decoder, the drag decoder employs the same three-branch feature extraction module, followed by a three-layer MLP that outputs the drag coefficient $C_d$.

\subsubsection{Training Strategy}
To achieve stable convergence and promote interaction between geometric and physical information, we adopt a two-stage training strategy consisting of modular pretraining followed by joint fine-tuning.

\noindent \textbf{Stage 1: Independent Training.}
For the encoder and shape decoder, we initialize from the pretrained weights of Dora~\cite{Chen25b} and fine-tune them on our dataset.
The training objective combines the SDF loss with a Kullback–Leibler (KL) regularization term:
\vspace{-3mm}
\begin{equation}
    \mathcal{L}_{\mathrm{shape}} = \lambda_{\mathrm{sdf}}\mathcal{L}_{\mathrm{sdf}} + \lambda_{\mathrm{KL}}\mathcal{L}_{\mathrm{KL}},
\vspace{-3mm}
\end{equation}
\begin{equation}
    \mathcal{L}_{\mathrm{sdf}} = \| s - \hat{s} \|_2^2,
\vspace{-2mm}
\end{equation}
where $s$ and $\hat{s}$ denote the predicted and ground-truth SDF values, respectively. With the encoder frozen, the pressure and drag decoders are trained separately from scratch using a composite loss that combines mean absolute error (MAE) and mean squared error (MSE):
\begin{align}
\mathcal{L}_{\mathrm{press}} &=
\| p - \hat{p} \|_1 + \| p - \hat{p} \|_2^2, \\
\mathcal{L}_{\mathrm{drag}} &=
\| C_d - \hat{C}_d \|_1 + \| C_d - \hat{C}_d \|_2^2,
\end{align}
where $\hat{p}$ and $\hat{C}_d$ represent the ground-truth surface pressure and drag coefficient, respectively.

\noindent \textbf{Stage 2: Joint Fine-Tuning.}
Finally, all components, including the encoder and three decoders, are jointly optimized under the combined objective:
\begin{equation}
\mathcal{L}_{\mathrm{total}} =
\lambda_{\mathrm{shape}}\mathcal{L}_{\mathrm{shape}} +
\lambda_{\mathrm{press}}\mathcal{L}_{\mathrm{press}} +
\lambda_{\mathrm{drag}}\mathcal{L}_{\mathrm{drag}},
\end{equation}
where $\lambda_{\mathrm{shape}}$, $\lambda_{\mathrm{press}}$, and $\lambda_{\mathrm{drag}}$ balance the contributions of different tasks.
This staged training ensures stable convergence and a physics-informed latent space, forming the basis for physics-guided generation (Sec.~\ref{sec:opt}).

\subsection{Physics-Guided Shape Generation}
\label{sec:opt}
Common physics-guided shape generation methods~\cite{Vatani25a, Zhan25a} iteratively update the geometry toward target physical objectives. However, without explicit awareness of the underlying shape manifold, these methods struggle to recover once the geometry becomes suboptimal or distorted.
In contrast, we integrate the flow-matching 3D shape generation and physics-based guidance into a unified framework, enforcing stable convergence toward geometrically plausible and physically valid results. To achieve this, we propose an alternating generation paradigm, in which we iteratively perform a velocity-based update with physics-based regularization and a physical refinement. Please refer to Alg.~\ref{algo:alternative} for the full algorithm description.

\subsubsection{Physics-Regularized Flow Matching}
\label{sub2sec:sampling}
We first train a flow matching model that learns a shape manifold over plausible geometries, which is used for generating high-fidelity 3D shapes, conditionally from images or unconditionally from noise. We adopt rectified flow~\cite{Liu23e} to formulate the flow matching model, which learns a velocity field that transports noise $\boldsymbol{\epsilon} \sim \mathcal{N}(0,1)$ toward data $\mathbf{z}_1$.
The forward process is expressed as a linear interpolation to obtain the noisy sample $\mathbf{z}_{t_n}$ in time step $t_n \in [0,1]$:
\begin{equation}
\label{eq:noisy_sample}
    \mathbf{z}_{t_n} = t_n \, \mathbf{z}_1 + (1-t_n) \, \boldsymbol{\epsilon} ,
\end{equation}
and the corresponding velocity field is defined as:
\begin{equation}
    \mathbf{u}_{t_n} = \frac{d\mathbf{z}_{t_n}}{dt_n} = \mathbf{z}_1 - \boldsymbol{\epsilon}.
\end{equation}
The model is trained to predict this velocity field from an optional condition $\mathbf{c} = 
\left\{
  \begin{array}{ll}
    \textbf{I}, & \text{if conditional on image} \\
    \varnothing, & \text{if unconditional}
  \end{array}
\right.$. The network architecture is provided in the supplementary material.
During the forward update, the reverse step is computed using the predicted velocity $\hat{\mathbf{u}}(\mathbf{z}_{t_n}, {t_n}, \mathbf{c})$ as:
\begin{equation}
\label{eq:reverse}
    \mathbf{z'}_{t_{n+1}} = \mathbf{z}_{t_{n}} - (t_{n+1} - t_{n})\hat{\mathbf{u}}(\mathbf{z}_{t_n}, {t_n}, \mathbf{c}).
\end{equation}
Inspired by classifier guidance~\cite{Dhariwal21a}, we incorporate a physics-based regularization term by using the drag decoder $\mathcal{D}_d$, trained in Sec.~\ref{sec:vae}, as a physics-aware estimator during sampling.
As shown in Fig.~\ref{fig:overview} (bottom left), at each time step $t_n$, $\mathcal{D}_d$ predicts the drag coefficient of $\mathbf{z}_{t_n}$, and the gradient of its deviation from the target $d_{\text{tar}}$ guides the update:
\begin{equation}
\mathbf{z}_{t_{n+1}} = \mathbf{z}'_{t_{n+1}} - \lambda_d \nabla_{\mathbf{z}_{t_n}} \big\| \mathcal{D}_d(\mathbf{z}_{t_n}) - d_{\text{tar}} \big\|_2^2.
\end{equation}
This physics-based regularization softly steers the flow trajectory toward physically plausible regions of the manifold, promoting the generation of physically valid shapes.

\begin{algorithm}[t]
\small
\caption{Physics-Guided Generation}
\KwIn{initial noise $\mathbf{z}_0$, shape decoder $\mathcal{D}_s$, drag decoder $\mathcal{D}_d$, pressure decoder $\mathcal{D}_p$, velocity estimator $\hat{\mathbf{u}}(\cdot)$, sampling steps $N$, refinement steps $M$, target drag coefficient $d_{\text{tar}}$, weights $\lambda_d, \lambda_x, \lambda_y, \lambda_z$,  normals $\mathbf{n}_x$, $\mathbf{n}_y$, $\mathbf{n}_z$, local face area $A$.}
$n_s = 0$, $t_n \leftarrow \frac{n}{N}$\;

\For{$k = 1$ \KwTo $K$}{
\hfill\textit{// Phase 1: Physics-Regularized Flow Matching}

\For{$n = n_s$ \KwTo $N-1$}{
    $\mathbf{z'}^k_{t_{n+1}} \leftarrow \mathbf{z}^k_{t_{n}} - (t_{n+1} - t_{n})\, \hat{\mathbf{u}}(\mathbf{z}^k_{t_{n}}, t_{n})$\;
    $\mathbf{z}^k_{t_{n+1}} \leftarrow \mathbf{z'}^k_{t_{n+1}} - \lambda_d\, \nabla_{\mathbf{z}^k_{t_n}} \big\| \mathcal{D}_d(\mathbf{z}^k_{t_n}) - d_{\text{tar}} \big\|^2_2$\;
}
$p \leftarrow \mathcal{D}_p(\mathbf{z}^k_1)$\; \vspace{-3.5mm} \hfill\textit{// Pressure Prediction}

\For{$m = 1$ \KwTo $M$}{ \vspace{-3.2mm} \hfill\textit{// Phase 2: Physical Refinement}
    $\displaystyle F_s = \sum_{i=1}^{V} p_i\, \mathbf{n}_{s,i}\, A_i,
\quad s \in \{x, y, z\} $\;
    $\displaystyle \mathcal{L} \leftarrow \lambda_x \|F_x\|_2 + \lambda_y \|F_y\|_2 + \lambda_z \mathrm{ReLU}(F_z) $\;
  $\mathbf z^{k}_{1, m} \leftarrow \mathbf z^{k}_{1, m-1} - \nabla_{\mathbf z^{k}_{1, m-1}}\, \mathcal L$\;
}
$\hat{\mathbf{z}}^k_1 = \mathbf z^{k}_{1, M}$\;
$n_s = \left\lfloor 0.75N \right\rfloor$\;
$\mathbf{z}^{k+1}_{t_{n_s}} = t_{n_s}\hat{\mathbf{z}}^k_1 + (1-t_{n_s})\mathbf{\epsilon}$\; \vspace{-3.5mm} \hfill\textit{// Re-noise}
}
\KwOut{$\mathbf{z}^{K}_1$}
\label{algo:alternative}
\end{algorithm}

\subsubsection{Physical Refinement}
\label{sub2sec:physrefine}
Given the clean latent $\mathbf{z}_1^k$ from the flow matching model at the $k$-th iteration, the shape and pressure decoders reconstruct the 3D geometry and its corresponding dense surface pressure.
This dense field provides localized physical information that enables fine-grained aerodynamic refinement. 
We compute directional forces using the surface pressure $p$, face normals $\mathbf{n}_{s}$, and face areas $A$, where $s \in \{x, y, z\}$ denotes drag, lateral, and lift directions, respectively:
\begin{equation}
    F_s = \sum_{i=1}^{V} p_i\, \mathbf{n}_{s,i}\, A_i,
\qquad s \in \{x, y, z\}.
\end{equation}
The corresponding physical losses are:
\begin{equation}
    \mathcal{L}_x = \|F_x\|_2,
\quad
\mathcal{L}_y = \|F_y\|_2,
\quad
\mathcal{L}_z = \mathrm{ReLU}(F_z),
\end{equation}
\begin{equation}
\mathcal{L} =
\lambda_{x}\mathcal{L}_{x} +
\lambda_{y}\mathcal{L}_{y} +
\lambda_{z}\mathcal{L}_{z},
\end{equation}
where $\mathcal{L}_x$ encourages minimal drag, $\mathcal{L}_y$ encourages lateral force symmetry, and $\mathcal{L}_z$ enforces negative lift for traction.
The gradients of $\mathcal{L}$ are then backpropagated to the latent code $\mathbf{z}_1^{k}$, yielding the refined $\hat{\mathbf{z}}_1^{k}$ for the next iteration.

\subsubsection{Alternating Update Strategy}
To ensure that generation satisfies both geometric plausibility and physical validity rather than drifting toward either extreme, we adopt an alternating update strategy that couples the velocity-based update in Sec.~\ref{sub2sec:sampling} with the physical refinement in Sec.~\ref{sub2sec:physrefine}. As shown in Alg.~\ref{algo:alternative}, at iteration $k$, the velocity-based update gradually denoises the latent code and encourages lower drag from time step $t_{n_s}$. Given the sampled latent $\mathbf{z}_1^{k}$, the physical refinement stage applies direction-aware physical gradients for $M$ steps. The refined latent $\hat{\mathbf{z}}_1^{k}$ is then re-noised to step $t_{n_s}$, following Eq.\ref{eq:noisy_sample} to produce $\mathbf{z}_{t_{n_s}}^{k+1}$, which initializes the next velocity-based update phase. After $K$ alternating iterations, the process converges to geometrically and physically plausible shapes.
\section{Experiments}
\label{sec:exp}

\subsection{Experimental Setup}

\noindent \textbf{Datasets.}
We conduct training and evaluation on the DrivAerNet++~\cite{Elrefaie24a} dataset, a vehicle aerodynamic benchmark featuring extensive geometric diversity and high-fidelity CFD simulations, including drag coefficients, surface pressure fields, and full 3D flow fields.
To evaluate generalization, we test our method on ShapeNet~\cite{Chang15} vehicles and further apply it to the structural optimization task~\cite{Zhan25a}.

\noindent \textbf{Evaluation Metrics.}
We evaluate our method on \textit{physics-aware shape generation} and \textit{aerodynamic property estimation}.
For shape generation, we use F-score~\cite{Chen25b}, Chamfer Distance (CD)~\cite{Chen25b}, accuracy~\cite{Chen25b}, and Intersection over Union (IoU)~\cite{Chen25b} to assess geometric fidelity.
For the estimation task, we adopt standard regression metrics including Mean Squared Error (MSE)~\cite{Chen25c}, Mean Absolute Error (MAE)~\cite{Chen25c}, Maximum Absolute Error (Max AE)~\cite{Chen25c}, Relative $L_2$ Error (Rel L2)~\cite{Chen25c}, and Relative $L_1$ Error (Rel L1)~\cite{Chen25c}.
To evaluate real-world performance, we conduct high-fidelity CFD simulations in \texttt{OpenFOAM}~\cite{Jasak07} to compute the drag coefficient $C_d$.
Detailed definitions are provided in the supplementary material.

\subsection{Physics-Guided Shape Generation}
\label{exp:single-view}

\begin{table}[t]
\footnotesize
\centering
\caption{{Unified generation vs. post-optimization.
“$\dagger$” denotes a stronger setting (500 steps, learning rate $10^{-3}$) than the conservative one (100 steps, $10^{-5}$). 
``O-Acc." denotes overall accuracy.
}
\vspace{-2mm}
}
\setlength{\tabcolsep}{0.9pt}
\renewcommand{\arraystretch}{1.1}
\begin{tabular}{l|cccccccc}
\toprule
\textbf{Model} & \textbf{F-score (0.01)$\times$100$\uparrow$} & \textbf{CD$\times$1000$\downarrow$} & \textbf{O-Acc.} \\
\midrule
Generation w/o phys.  & 74.03  &  27.14 &  60.86  \\ 
SP-VAE+TripOptimizer~\cite{Vatani25a} & 73.93  & 27.13  & 60.89   \\ 
SP-VAE+TripOptimizer$\dagger$~\cite{Vatani25a} &  67.70 &  32.78 &  58.75    \\ 
\midrule
Ours & \textbf{89.65}  &  \textbf{20.99} &  \textbf{66.48}    \\
\bottomrule
\end{tabular}
\label{tab:opt_comp}
\vspace{-3mm}
\end{table}

\begin{figure}
    \centering
    \includegraphics[width=\linewidth]{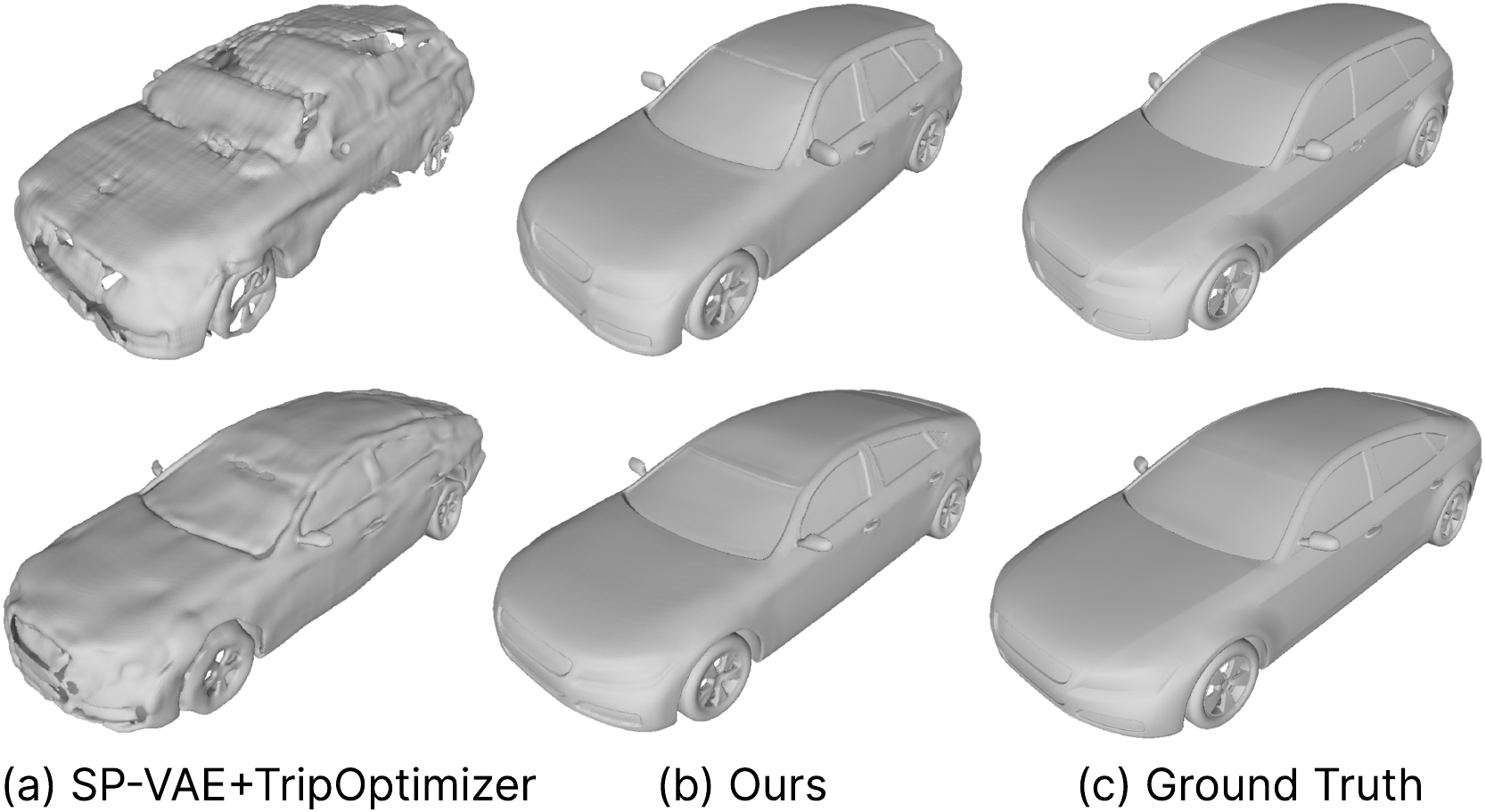}
    \vspace{-5mm}
    \caption{Qualitative comparison of post-optimization and our unified generation. SP-VAE + TripOptimizer produces distorted shapes and fails to recover them, whereas our alternating method restores plausible surfaces closer to the ground truth.
    }
    \label{fig:opt_comp}
\vspace{-2mm}
\end{figure}

\noindent \textbf{Unified Generation vs. Post-Optimization.}
Starting from a physically imperfect initial shape, TripOptimizer~\cite{Vatani25a} predicts the drag coefficient of the current shape and iteratively updates the shape by minimizing its discrepancy from the target value. This pipeline treats generation and physics-based refinement as two separate stages, whereas our method integrates them into a unified framework. Table~\ref{tab:opt_comp} evaluates physics-aware generation accuracy by measuring the similarity between generated shapes and the ground truth. Since TripOptimizer is not publicly available, we reproduce its two-stage strategy using our SP-VAE. Under a conservative setting (100 steps, learning rate $= 10^{-5}$), the geometry changes only marginally, whereas a stronger setting (500 steps, learning rate $= 10^{-3}$) causes severe deformations and lower accuracy. As shown in Fig.~\ref{fig:opt_comp}, once the shape is distorted, the two-stage method cannot recover it because it lacks awareness of the shape manifold. In contrast, our alternating strategy jointly promotes shape plausibility and physical efficiency, correcting distortions and producing more accurate shapes.

\begin{figure}
    \centering
    \includegraphics[width=\linewidth]{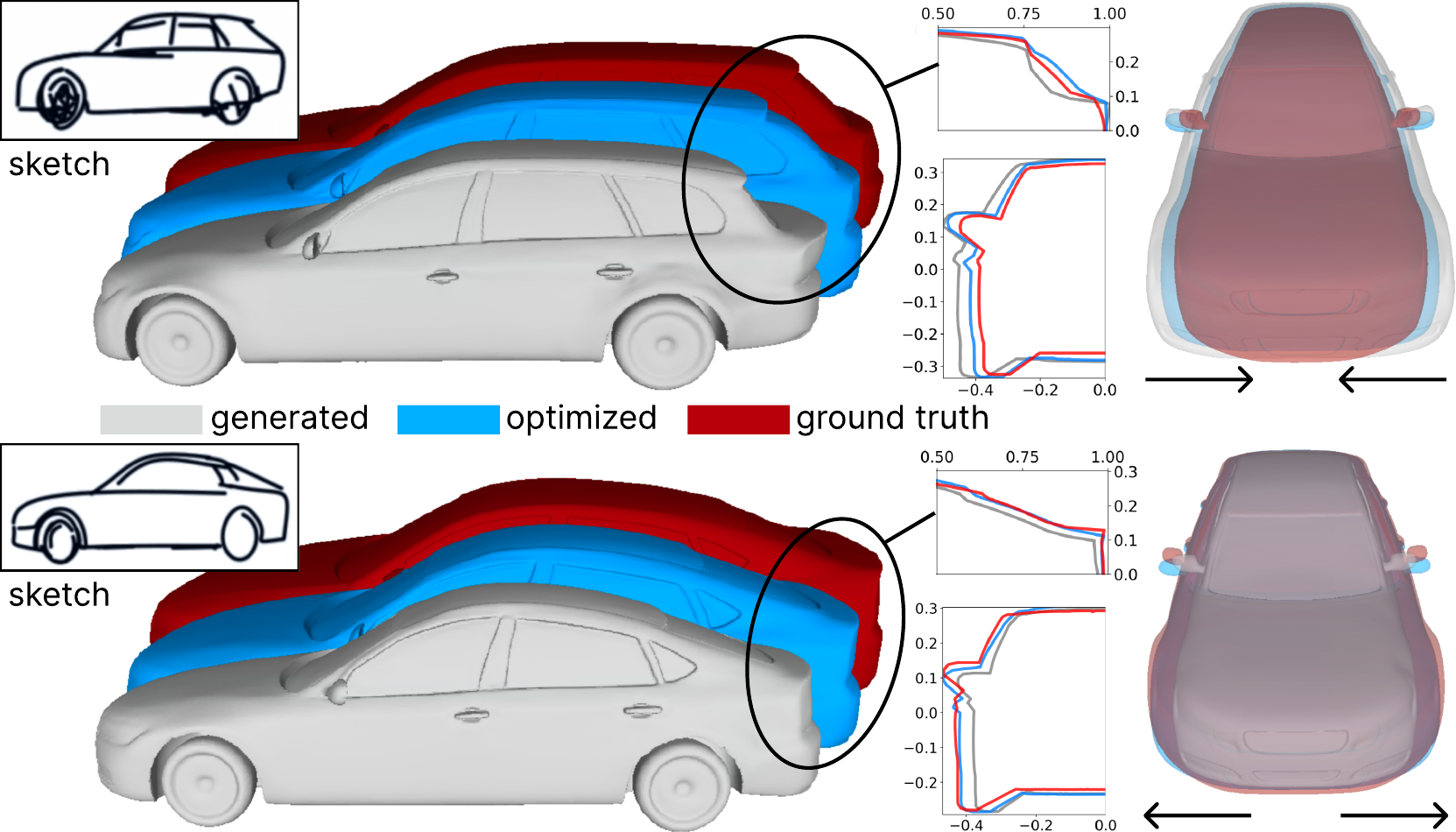}
    \vspace{-7mm}
    \caption{Physical information improves shape generation accuracy. Starting from the physically unguided generation (gray), refining toward the target drag coefficient (blue) aligns the geometry more closely with the ground-truth shape (red).}
    \vspace{-2mm}
    \label{fig:exp_fixdrag}
\end{figure}

\noindent \textbf{Shape Generation under Target Drag.}
As shown in Fig.~\ref{fig:exp_fixdrag}, starting from the generated shape from the sketch image without physical information (gray), refining toward the target drag coefficient (blue) aligns the geometry more closely with the ground-truth shape (red), particularly in the rear, roof and vehicle width, where aerodynamic effects are most sensitive.
Quantitatively, Table~\ref{tab:fixdrag} shows that applying the target drag coefficient improves the F-score (0.01) by {21.09\%} and reduces the Chamfer Distance (CD) by {22.68\%} compared to the unguided generation, indicating higher geometric accuracy.
These results demonstrate that a target drag coefficient provides additional cues that help alleviate the depth ambiguity inherent in 2D-to-3D generation, leading to improved 3D shape.

\begin{table}[t]
\footnotesize
\centering
\caption{Shape accuracy under target drag coefficient $d_{tar}$.}
\vspace{-2mm}
\setlength{\tabcolsep}{9.6pt}
\renewcommand{\arraystretch}{1.1}
\begin{tabular}{l| c | c }
\toprule
\textbf{Shape} & \textbf{F-score (0.01) $\times 100$} $\uparrow$ & \textbf{CD $\times 1000$} $\downarrow$ \\
\midrule
w/o target $d_{tar}$ & 74.03 & 27.14 \\
w/ ~~target $d_{tar}$  & 89.65 ({\textbf{21.09\%} $\uparrow$}) & 20.99 ({\textbf{22.68\%} $\uparrow$}) \\
\bottomrule
\end{tabular}
\label{tab:fixdrag}
\vspace{-4mm}
\end{table}

\begin{table}[t]
\footnotesize
\centering
\caption{Comparison of shapes from a real image without (w/o phys.) and with (w/ phys.) physical guidance.}
\vspace{-2mm}
\setlength{\tabcolsep}{8.8pt}
\begin{tabular}{c|l|cc}
\toprule
\textbf{Case} & \textbf{Metric} & \textbf{w/o phys.} & \textbf{w/ phys.} \\
\midrule
\multirow{3}{*}{\rotatebox{90}{Case a}} 
 & Chamfer Distance ($\times 10^4$) $\downarrow$ & 20.98 & \textbf{2.38} \\
 & Normal Consistency $\uparrow$ & 0.78 & \textbf{0.90} \\
 & F-score (0.01) $\uparrow$ & 0.41 & \textbf{0.85} \\
\midrule
\multirow{3}{*}{\rotatebox{90}{Case b}} 
 & Chamfer Distance ($\times 10^4$) $\downarrow$ & 8.06 & \textbf{0.83} \\
 & Normal Consistency $\uparrow$ & 0.88 & \textbf{0.96} \\
 & F-score (0.01) $\uparrow$ & 0.59 & \textbf{0.98} \\
\bottomrule
\end{tabular}
\label{tab:real_image}
\vspace{-2mm}
\end{table}

\begin{figure}
    \centering
    \includegraphics[width=\linewidth]{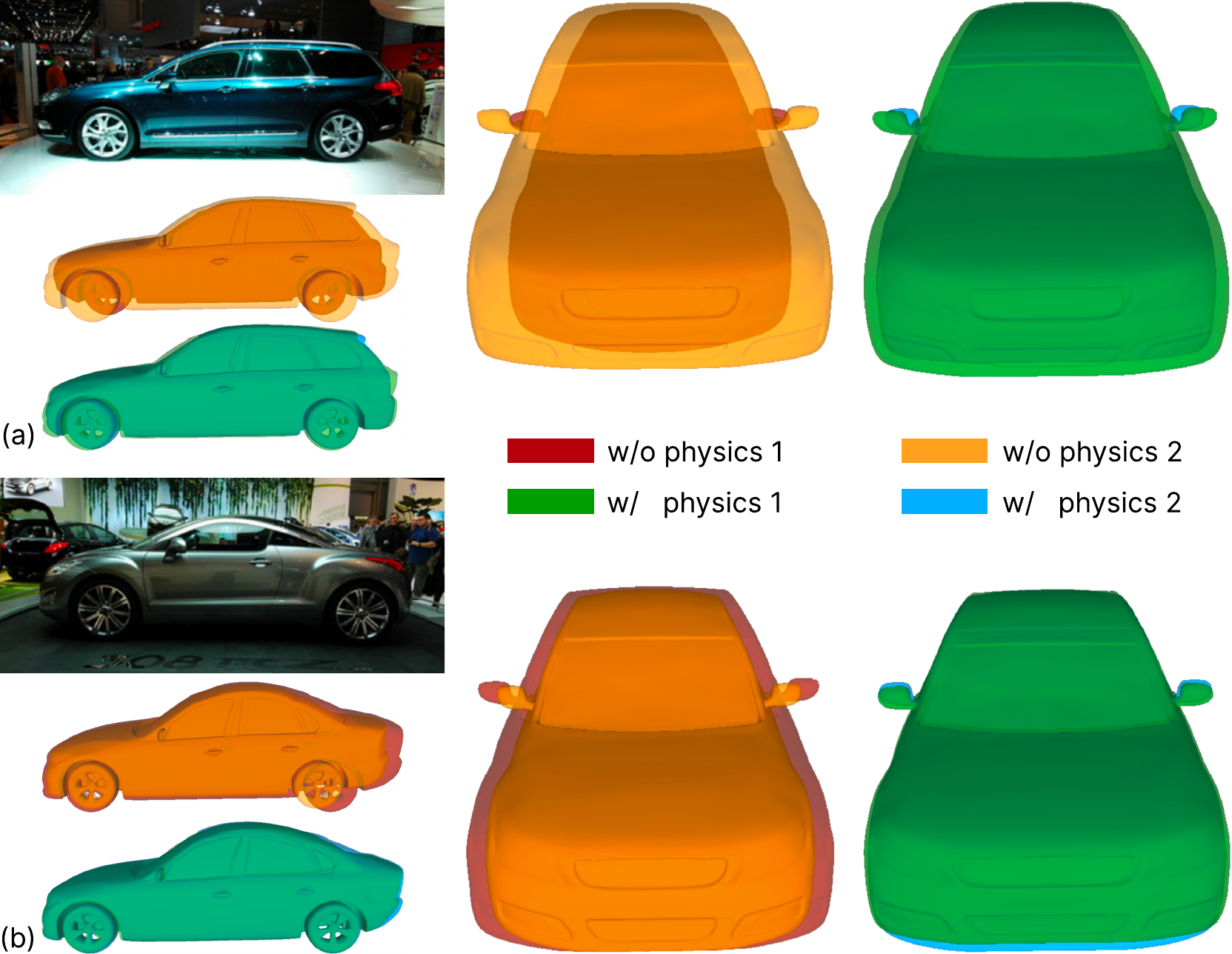}
\vspace{-4mm}
    \caption{Physical information mitigates depth ambiguity in 3D generation from a real single-view image. Two shapes are first generated from the same image using different initial noises (red and orange cars). When physical guidance is applied, the resulting shapes (blue and green cars) converge to similar front-view widths.
    }
    \label{fig:real_car}
\vspace{-1mm}
\end{figure}

\noindent \textbf{Single-View Real Image with Physical information.}
We evaluate our method on a real single-view image using cross-evaluation, as ground-truth meshes are unavailable.
As shown in Fig.~\ref{fig:real_car}, without physical information, two shapes generated from the same image but different initial noises (red and orange cars) share similar side views yet differ in front-view width due to depth ambiguity. When physical guidance, which enforces an approximate target drag coefficient and minimizes directional forces, is applied (blue and green cars), the generated 3D shapes converge to similar front-view widths.
Table~\ref{tab:real_image} quantitatively shows reduced inter-shape discrepancy, indicating that physical guidance effectively mitigates depth ambiguity.

\begin{table}[t]
\footnotesize
\centering
\caption{Comparison on shape reconstruction. ``$\dagger$" indicates fine-tuning on the DrivAerNet++~\cite{Elrefaie24} dataset. ``O-", ``S-", and ``C-" denote overall, sharp, and coarse, respectively.}
\vspace{-2mm}
\setlength{\tabcolsep}{2.8pt}
\renewcommand{\arraystretch}{1.1}
\begin{tabular}{l| c c c c c c}
\toprule
\textbf{Model} & \textbf{O-Acc.} & \textbf{O-IoU} & \textbf{S-Acc.} & \textbf{S-IoU}  & \textbf{C-Acc.} & \textbf{C-IoU}  \\
\midrule
3DSet2Vector~\cite{Zhang23d} & 73.58 & 51.28 & 64.61 & 49.80 & 83.17 & 54.32 \\ 
Hunyuan3D 2.1~\cite{hunyuan3d2025hunyuan3d} & 89.43 & 76.55 & 87.19 & 77.57 & 91.81 & 74.62 \\
Hi3DGen~\cite{ye2025hi3dgen}       & 91.47 & 81.52 & 89.37 & 82.76 & 93.67 & 80.08\\
Dora~\cite{Chen25b} & 86.49 & 71.28 & 84.78 & 74.16 & 88.32 & 66.04 \\ 
Dora$\dagger$~\cite{Chen25b} & 95.31 & 88.61  & 94.32 & 89.13 & 96.37 & 87.62 \\
\midrule
Ours & \textbf{96.73} & \textbf{91.89} & \textbf{95.64} & \textbf{91.50} & \textbf{97.89} & \textbf{92.63}\\
\bottomrule
\end{tabular}
\label{tab:shape_est}
\vspace{-3mm}
\end{table}
\subsection{Shape Reconstruction and Physics Estimation}

\noindent \textbf{Comparison of Shape Reconstruction.} Here, we use our trained encoder and shape decoder in SP-VAE to reconstruct 3D meshes from the input point clouds. The results in Table~\ref{tab:shape_est} show that our model outperforms 3DSet2Vector~\cite{Zhang23d}, the VAEs of Hunyuan3D 2.1~\cite{hunyuan3d2025hunyuan3d} and Hi3DGen~\cite{ye2025hi3dgen}, pure Dora~\cite{Chen25b} model, and its fine-tuned version on the DrivAerNet++~\cite{Elrefaie24a} dataset, across all evaluation metrics. 
The improvements highlight that our physics-informed representations contribute to more high-fidelity shape reconstructions.

\begin{table}[t]
\footnotesize
\centering
\caption{Performance comparison on drag coefficient estimation.
}
\vspace{-3mm}
\setlength{\tabcolsep}{10.2pt}
\renewcommand{\arraystretch}{1.1}
\begin{tabular}{l| c c c}
\toprule
\multirow{2}{*}{\textbf{Model}} & \textbf{MSE$\downarrow$} & \textbf{MAE$\downarrow$}& \textbf{Max AE$\downarrow$}\\
 & $(\times 10^{-5})$ & $(\times 10^{-3})$ & $(\times 10^{-2})$\\
\midrule
GCNN~\cite{Jiang19c} & 17.1 & 10.43 & 15.03 \\
RegDGCNN~\cite{Elrefaie24a} & 14.2 & 9.31 & 12.79 \\
PointNet~\cite{Qi17c} & 14.9 & 9.60 & 12.45 \\ 
TripNet~\cite{Chen25c} & 9.1 & 7.17 & 7.70 \\ 
\midrule
Ours & \textbf{4.0} & \textbf{4.83} & \textbf{2.70}\\
\bottomrule
\end{tabular}
\label{tab:drag_est}
\vspace{-2mm}
\end{table}
\begin{table}[t]
\footnotesize
\centering
\caption{Performance comparison on pressure field prediction.
}
\vspace{-3mm}
\setlength{\tabcolsep}{5.6pt}
\renewcommand{\arraystretch}{1.1}
\begin{tabular}{l| c c c c}
\toprule
\multirow{2}{*}{\textbf{Model}} & \textbf{MSE$\downarrow$} & \textbf{MAE$\downarrow$}  & \textbf{Rel L2}$\downarrow$ & \textbf{Rel L1}$\downarrow$ \\
 & $(\times 10^{-2})$ & $(\times 10^{-1})$  & $(\%)$ & $(\%)$ \\
\midrule
RegDGCNN~\cite{Elrefaie24a} & 8.29 & 1.61 & 27.72 & 26.21 \\ 
FigConvNet~\cite{Choy25a} & \underline{4.99} & \underline{1.22} & 20.86 & 21.12 \\ 
Transolver~\cite{Wu24a} & 7.15 & 1.41 & 23.87 & 22.57 \\ 
TripNet~\cite{Chen25c} & 5.14 & 1.25 & \underline{20.05} & 20.93 \\ 
\midrule
Ours & \textbf{4.55} & \textbf{1.09} & \textbf{20.02} & \textbf{17.78} \\
\bottomrule
\end{tabular}
\label{tab:pressure_est}
\vspace{-2mm}
\end{table}

\noindent \textbf{Comparison of Physics Estimation.} We predict the physical properties from the point cloud, utilizing the pressure decoder and drag decoder. 
As shown in Tables~\ref{tab:drag_est} and~\ref{tab:pressure_est}, our method achieves the best performance in both drag coefficient prediction and surface pressure estimation, whereas the baselines estimate physical quantities solely from shape inputs without leveraging complementary information. These results show that the joint shape-physics latent representation captures the correlation between geometry and aerodynamics, improving both shape reconstruction and physics estimation.

\begin{table*}[t]
\footnotesize
\centering
\caption{Ablation study on the training strategy of SP-VAE. ``O-", ``S-", and ``C-" denote overall, sharp, and coarse, respectively.}
\vspace{-2mm}
\setlength{\tabcolsep}{3pt}
\renewcommand{\arraystretch}{1.1}
\begin{tabular}{l| c c c | c c c c | c c c c c c}
\toprule
\multirow{3}{*}{\textbf{Model}} 
& \multicolumn{3}{|c}{\textbf{Drag Estimation}} 
& \multicolumn{4}{|c}{\textbf{Surface Pressure Estimation}} 
& \multicolumn{6}{|c}{\textbf{Shape Reconstruction}} \\
& {MSE$\downarrow$} & {MAE$\downarrow$}& {Max AE$\downarrow$} & {MSE$\downarrow$} & {MAE$\downarrow$}  & {Rel L2}$\downarrow$ & {Rel L1}$\downarrow$ & {O-Acc.} & {O-IoU} & {S-Acc.} & {S-IoU}  & {C-Acc.} & {C-IoU}  \\
& $(\times 10^{-5})$ & $(\times 10^{-3})$ & $(\times 10^{-2})$ & $(\times 10^{-2})$ & $(\times 10^{-1})$  & $(\%)$ & $(\%)$  & $(\%)$ & $(\%)$ & $(\%)$ & $(\%)$ & $(\%)$ & $(\%)$ \\
\midrule
Independent Training & 4.6 & 5.14 & 3.08 & 4.59 & \textbf{1.09} & 20.12 & 17.81 & 95.31 & 88.61  & 94.32 & 89.13 & 96.37 & 87.62\\ 
Joint Fine-tuning & \textbf{4.0} & \textbf{4.83} & \textbf{2.70} & \textbf{4.55} & \textbf{1.09} & \textbf{20.02} & \textbf{17.78}  & \textbf{96.73} & \textbf{91.89} & \textbf{95.64} & \textbf{91.50} & \textbf{97.89} & \textbf{92.63}\\
\bottomrule
\end{tabular}
\label{tab:ablation_finetune}
\end{table*}
\begin{figure*}
    \centering
    \includegraphics[width=\linewidth]{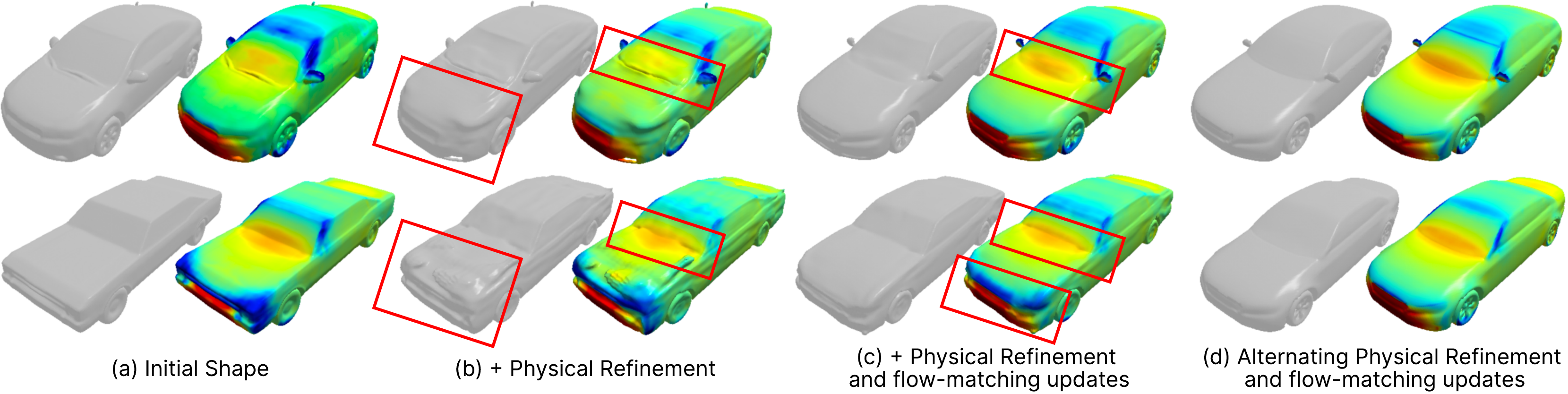}
    \vspace{-5mm}
    \caption{Visualization of generation with physical refinement and flow-matching updates, shown in terms of mesh geometry and surface pressure. Starting from a physically imperfect initialization (a), physical refinement improves physical objectives but introduces distortions (b). Adding flow-matching updates restore geometric plausibility but lead to non-uniform pressure (c). Alternating the two produces refined geometry and more uniform pressure, improving both visual quality and aerodynamic performance (d).}
    \vspace{-1mm}
    \label{fig:exp_alternating}
\end{figure*}

\subsection{Ablation Studies}
\noindent \textbf{Effectiveness of SP-VAE Training Strategy.}
As shown in Table~\ref{tab:ablation_finetune}, joint finetuning consistently improves performance across drag estimation, surface pressure prediction, and shape reconstruction.  
The most notable gain appears in {shape reconstruction}, where overall accuracy and IoU improve to {96.73} and {91.89}, respectively.  
These results confirm that joint finetuning fosters mutual reinforcement between geometric and physical representations, leading to a more coherent and expressive latent space.

\noindent \textbf{Effectiveness of Alternating Strategy.}
We evaluate joint shape quality and surface pressure on the out-of-distribution ShapeNet~\cite{Chang15}. As shown in Fig.~\ref{fig:exp_alternating}, \emph{physics refinement only} moves the initial imperfect shape toward the physical objectives. However, it introduces geometric artifacts without awareness of the shape manifold. Incorporating \emph{flow-matching updates} restores geometric plausibility, yet the physics objectives remain suboptimal. By iteratively \emph{alternating} between physics refinement and flow-matching updates, the method reconciles physical satisfaction with shape plausibility, achieving both aesthetic quality and aerodynamic performance.

\begin{figure}
    \centering
    \includegraphics[width=\linewidth]{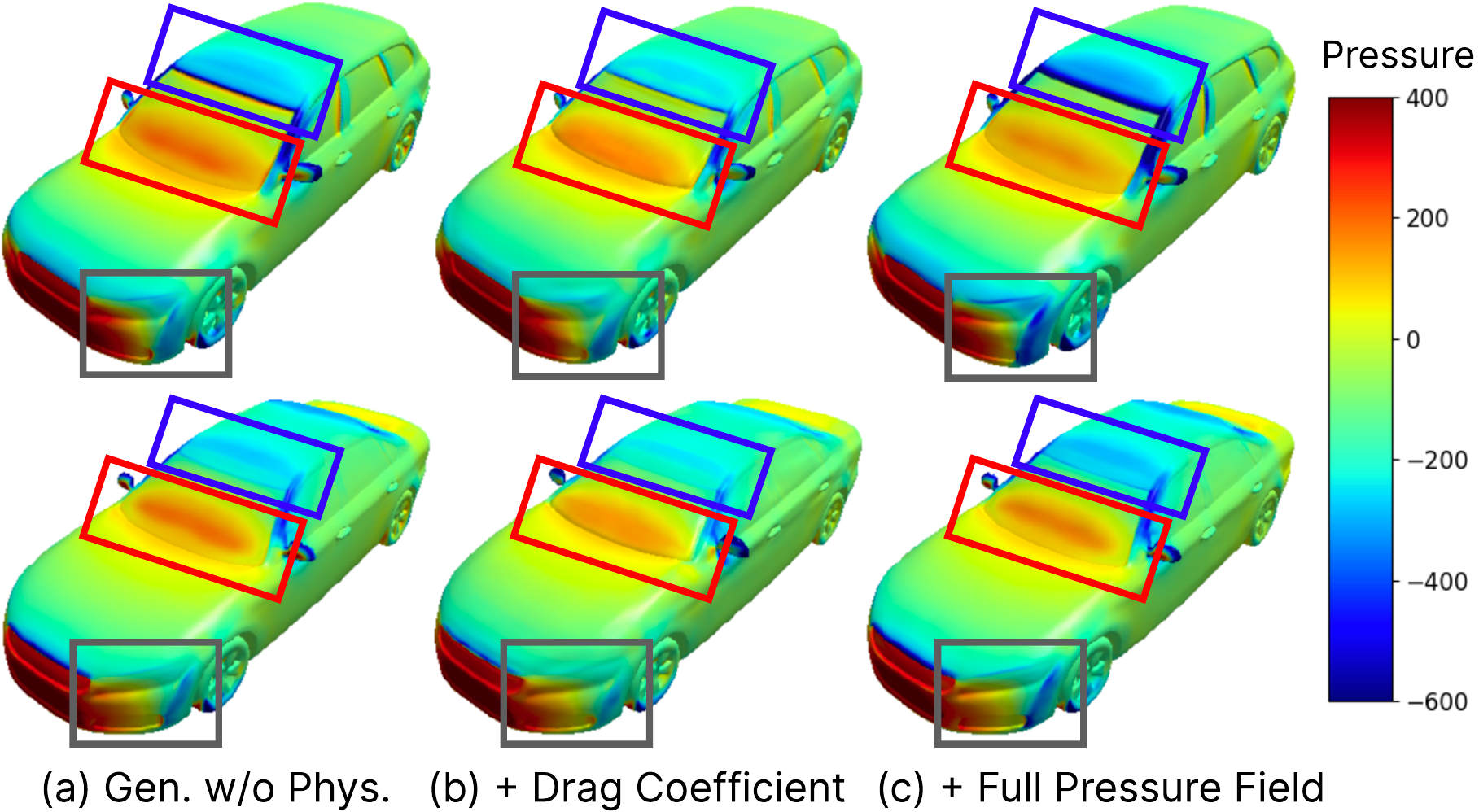}
    \vspace{-6mm}
    \caption{Surface pressure distribution of the initial shape and the generated results under drag coefficient or full pressure field.}
    \label{fig:exp_sparsedense}
    \vspace{-2mm}
\end{figure}
\noindent \textbf{Impact of Drag Coefficient and Full Pressure Field.}
Fig.~\ref{fig:exp_sparsedense} compares surface pressure distributions for shapes generated without physical guidance, with drag-only guidance, and with full pressure-field guidance.
Without physical guidance, the generated shape exhibits strong pressure over the hood and windshield (red box), unstable flow at the front-roof transition (blue box), and weak low-pressure continuity at the front corner (gray box), resulting in high drag.
Drag-only guidance reduces the front pressure peak (red box) and improves flow attachment, but the pressure distribution remains coarse and lacks local smoothness (red and blue boxes).
Full pressure-field guidance further suppresses high-pressure regions and yields smoother pressure distributions, thus enhancing aerodynamic performance.

\begin{table}[t]
\footnotesize
\centering
\caption{Ablation study on the components of physical decoder.}
\vspace{-2mm}
\setlength{\tabcolsep}{3.5pt}
\renewcommand{\arraystretch}{1.1}
\begin{tabular}{c c c| c c c c}
\toprule
\multirow{2}{*}{\textbf{Attn.}} & \multirow{2}{*}{\textbf{Channel}} & \multirow{2}{*}{\textbf{MLP}} & \textbf{MSE$\downarrow$} & \textbf{MAE$\downarrow$}  & \textbf{Rel L2} $\downarrow$ & \textbf{Rel L1} $\downarrow$ \\
& & & $(\times 10^{-2})$ & $(\times 10^{-1})$  & $(\%)$ & $(\%)$ \\
\midrule
& & $\checkmark$& 8.26 & 1.52 & 27.44 & 24.68 \\
& $\checkmark$& & 5.43 & 1.23 & 22.09 & 20.07 \\ 
$\checkmark$& & & 5.90 & 1.27 & 22.84 & 20.73 \\
\midrule
$\checkmark$ & $\checkmark$& & 5.20 & 1.21 & 21.49 & 19.63 \\
$\checkmark$& & $\checkmark$& 5.16 & 1.23 & 21.42 & 20.04 \\
& $\checkmark$& $\checkmark$& 5.15 & 1.21 & 21.47 & 19.62 \\
\midrule
$\checkmark$& $\checkmark$ & $\checkmark$& \textbf{4.59} & \textbf{1.09} & \textbf{20.12} & \textbf{17.81} \\
\bottomrule
\end{tabular}
\label{tab:ablation_physdec}
\vspace{-5mm}
\end{table}
\noindent \textbf{Impact of Physical Decoder Structure.}
Table~\ref{tab:ablation_physdec} analyzes the contribution of the attention, channel, and MLP branches in the proposed physics decoder.
Each pair of branches provides complementary benefits, and integrating all three achieves the best performance ({4.59} MSE, {1.09} MAE), indicating that multi-dimensional feature fusion effectively enhances physics prediction accuracy.

\begin{figure}[t]
    \centering
    \includegraphics[width=\linewidth]{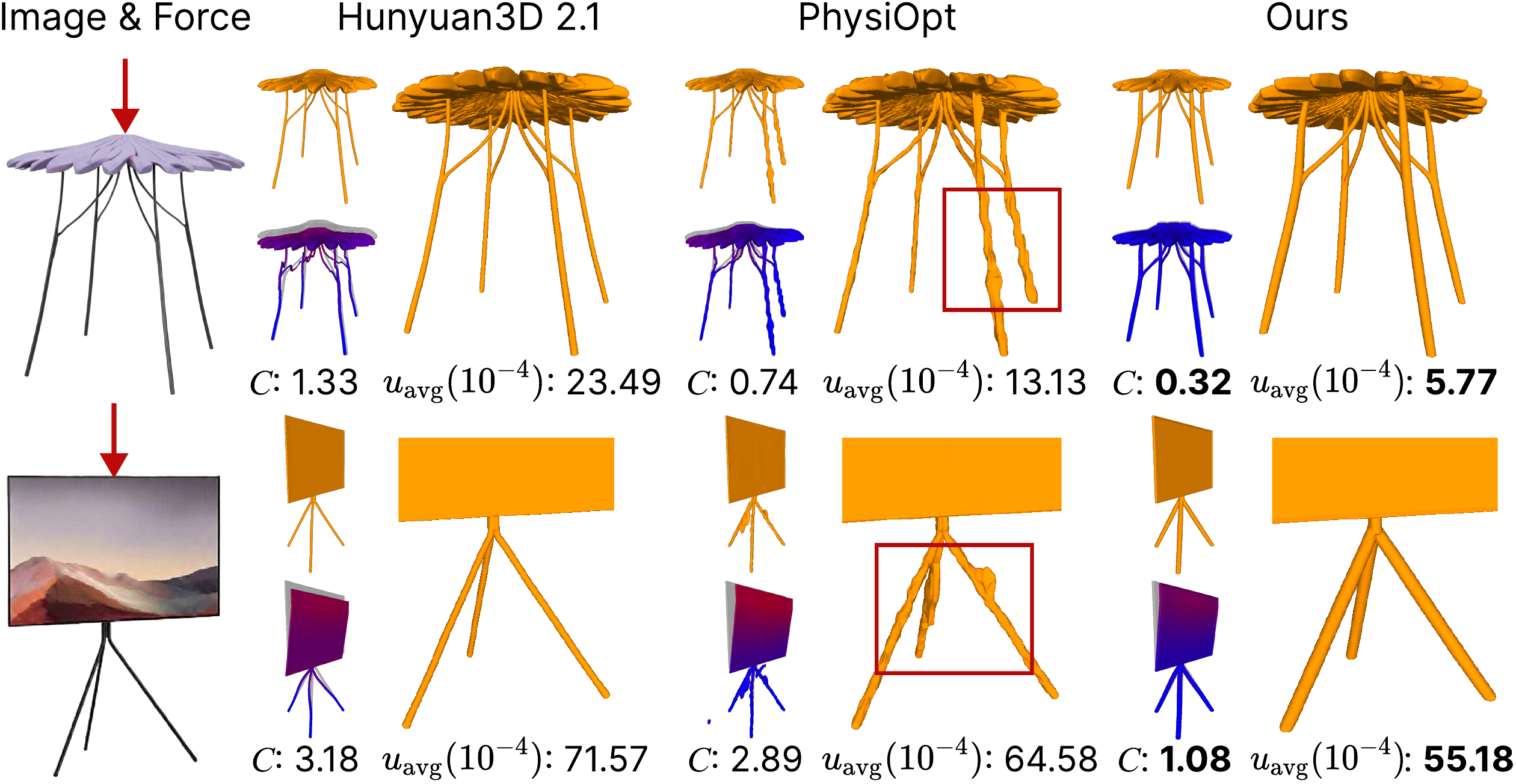}
    \vspace{-3.5mm}
    \caption{Comparisons for structural optimization. $C$: compliance (strain energy; lower is stiffer). $u_{\text{avg}}$: mean displacement.}
    \vspace{-5mm}
    \label{fig:force}
\end{figure}
\subsection{Generalization to Structural Optimization}
We further extend our physics-guided shape generation framework to structural optimization, where the goal is to minimize compliance under prescribed loads and boundary conditions, following PhysiOpt~\cite{Zhan25a}. As shown in Fig.~\ref{fig:force}, shapes generated by Hunyuan3D 2.1~\cite{hunyuan3d2025hunyuan3d} often contain thin legs that deform severely under load. PhysiOpt uses a physics simulator for optimization, but without shape-manifold awareness, it often distorts shapes and introduces artifacts. In contrast, our method complements physics with shape manifold, improving physical performance and shape quality. See the supplementary material for more details.
\section{Conclusion}
\label{sec:conclusion}
In this work, we present a flow matching paradigm with explicit physical guidance, including an alternating process between a velocity-based update with physics-aware regularization and a physical refinement. To bridge geometry and physics, we develop SP-VAE, which encodes both in a shared latent space, capturing their correlations and forming the basis for physics-guided generation. Experiments demonstrate that our method produces both physically efficient and aesthetically appealing shapes. It also enhances single-view reconstruction and aerodynamic performance, and generalizes to structural optimization. We hope this work inspires further research on physics-grounded generative modeling for realistic and functional 3D shapes.
\section*{Acknowledgements}
This work was supported in part by the Swiss National Science Foundation.

{
    \small
    \bibliographystyle{ieeenat_fullname}
    \bibliography{bib/short, bib/string, bib/ref}
}
\renewcommand{\thetable}{\Alph{table}}
\renewcommand{\theequation}{\Alph{equation}}
\renewcommand{\thefigure}{\Alph{figure}}
\renewcommand{\thesection}{\Alph{section}}
\renewcommand{\thesubsection}{\Alph{section}.\arabic{subsection}}

\clearpage
\setcounter{page}{1}
\setcounter{table}{0}
\setcounter{equation}{0}
\setcounter{figure}{0}
\setcounter{section}{0}
\setcounter{subsection}{0}
\maketitlesupplementary


\noindent This supplemental material includes the following sections:

(A) Implementation details.

(B) Additional experiments.

(C) Network architectures.

(D) Dataset details.

(E) Evaluation metrics.

(F) CFD simulation setup in OpenFOAM~\cite{Jasak07}.

(G) Generalization to structural optimization.

\section{Implementation Details}
\subsection{SP-VAE}
For each 3D mesh used for training, we extract $32,768$ uniform surface points $\mathbf{P}_u$ and $32,768$ salient edge points $\mathbf{P}_s$ using the Sharp Edge Sampling (SES) strategy~\cite{Chen25b}. As shown in Fig.~\ref{fig:sup_vae}, queries $\mathbf{Q}$ for cross-attention are constructed by applying Farthest Point Sampling (FPS)~\cite{Moenning03a} to $\mathbf{P}_u$ and $\mathbf{P}_s$, each downsampled to $1024$ points and concatenated into a $2048$-point set. We supervise the shape encoder-decoder using both uniformly sampled coarse points in the bounding box and sharp points perturbed around the ground-truth surface, with loss weights $\lambda_{\text{SDF}}=1$ and $\lambda_{\text{KL}}=0.001$. The shape encoder-decoder is trained for $1000$ epochs on $4\times$H100 GPUs (about 2 days). The pressure decoder is trained using pressure values sampled near the surface for $1500$ epochs on $4\times$H100 GPUs (about 21 hours). The drag decoder, which predicts a single global drag coefficient, is trained for $1500$ epochs and completes within one day on a single H100 GPU. After individual training, we perform joint fine-tuning for $500$ epochs on $4\times$H100 GPUs (about 15 hours) using a combined loss with $\lambda_{\text{shape}}=10$, $\lambda_{\text{physics}}=0.1$, and $\lambda_{\text{drag}}=10$. All experiments use $5819$ training samples and $1147$ test samples from DrivAerNet++~\cite{Elrefaie24}.

\subsection{Physics-Guided Shape Generation}
For flow-based generation, we adopt the rectified flow formulation~\cite{Liu23e}, using $100$ sampling steps at inference. In physics-guided shape generation, we incorporate a physics-based regularization term with weight $\lambda_d=0.03$ for drag guidance during velocity-based updates, while directional weights $\lambda_x=0.2$, $\lambda_y=0.1$, $\lambda_z=0.1$ are applied during the physical refinement phase. For alternating generation, we perform $K=20$ alternating iterations. In each iteration $k$,  we first apply $20$ steps of physical refinement to obtain the refined latent $\hat{\mathbf{z}}_1^{k}$, then re-noise it back to timestep $t_{n_s}=0.75$ to produce $\mathbf{z}_{t_{n_s}}^{k+1}$, which initializes the next velocity-based update phase. The full set of iterations takes roughly $210$ seconds. Overall, the procedure iteratively alternates between velocity-based updates and physics-based refinement, with each stage performing only a small number of steps (25 steps for velocity updates and 20 steps for physical refinement), gradually converging toward shapes that satisfy both geometric plausibility and physical efficiency.

\begin{table}[t]
\small
\centering
\caption{
Shape generation toward minimizing the drag coefficient. 
Image-unconditional generation (Unc.) minimizes drag without image, while conditional generation (Cond.) minimizes drag with image conditioning. Average drag coefficients simulated by OpenFOAM indicate aerodynamic performance. (SN: ShapeNet, DAN+: DrivAerNet++.)
}
\setlength{\tabcolsep}{4.5pt}
\renewcommand{\arraystretch}{1.1}
\begin{tabular}{l|c|c|c}
\toprule
\multirow{2}{*}{\textbf{Shape}} & \multicolumn{3}{c}{\textbf{Average Drag Coefficient}} \\
& \textbf{SN} (Unc.) &
\textbf{DAN+} (Unc.) &
\textbf{DAN+} (Cond.) \\
\midrule
w/o minimizing & 0.393 & 0.324 & 0.334 \\
w/ ~~minimizing & 0.304 & 0.274 & 0.312 \\
\midrule
\textbf{Improvement} &
{\textbf{22.70\%} $\uparrow$} &
{\textbf{15.47\%} $\uparrow$} &
{\textbf{6.53\%} $\uparrow$} \\
\bottomrule
\end{tabular}
\label{tab:min_opt}
\end{table}
\section{Additional Experiments}
\noindent \textbf{Shape Generation toward Minimizing Drag.}
Beyond leveraging a known target drag coefficient to improve reconstruction accuracy, our framework can also enhance aerodynamic performance by minimizing the drag coefficient.
Table~\ref{tab:min_opt} reports average results over $20$ samples per dataset, each simulated using OpenFOAM~\cite{Jasak07} (see Sec.~\ref{sec:sup_openfoam} for simulation details), covering both the in-distribution DrivAerNet++~\cite{Elrefaie24} dataset and the out-of-distribution ShapeNet~\cite{Chang15} car set.
Despite never observing ShapeNet geometries during training, our method achieves a substantial {22.7\%} drag reduction, demonstrating the generalization and the ability to maintain shape plausibility while improving aerodynamic performance. On DrivAerNet++, unconditional physics-guided generation (minimizing drag without image conditioning) reduces drag by {15.47\%}, whereas the conditional setting (minimizing drag with image conditioning) achieves a {6.53\%} reduction, as it balances aesthetic alignment with physical efficiency. These findings confirm that alternating prior- and physics-guided generation generalizes robustly to unseen geometries and improves aerodynamic performance while maintaining visual alignment when a conditional image is provided.

\section{Network Architectures}
\begin{figure}
    \centering
    \includegraphics[width=\linewidth]{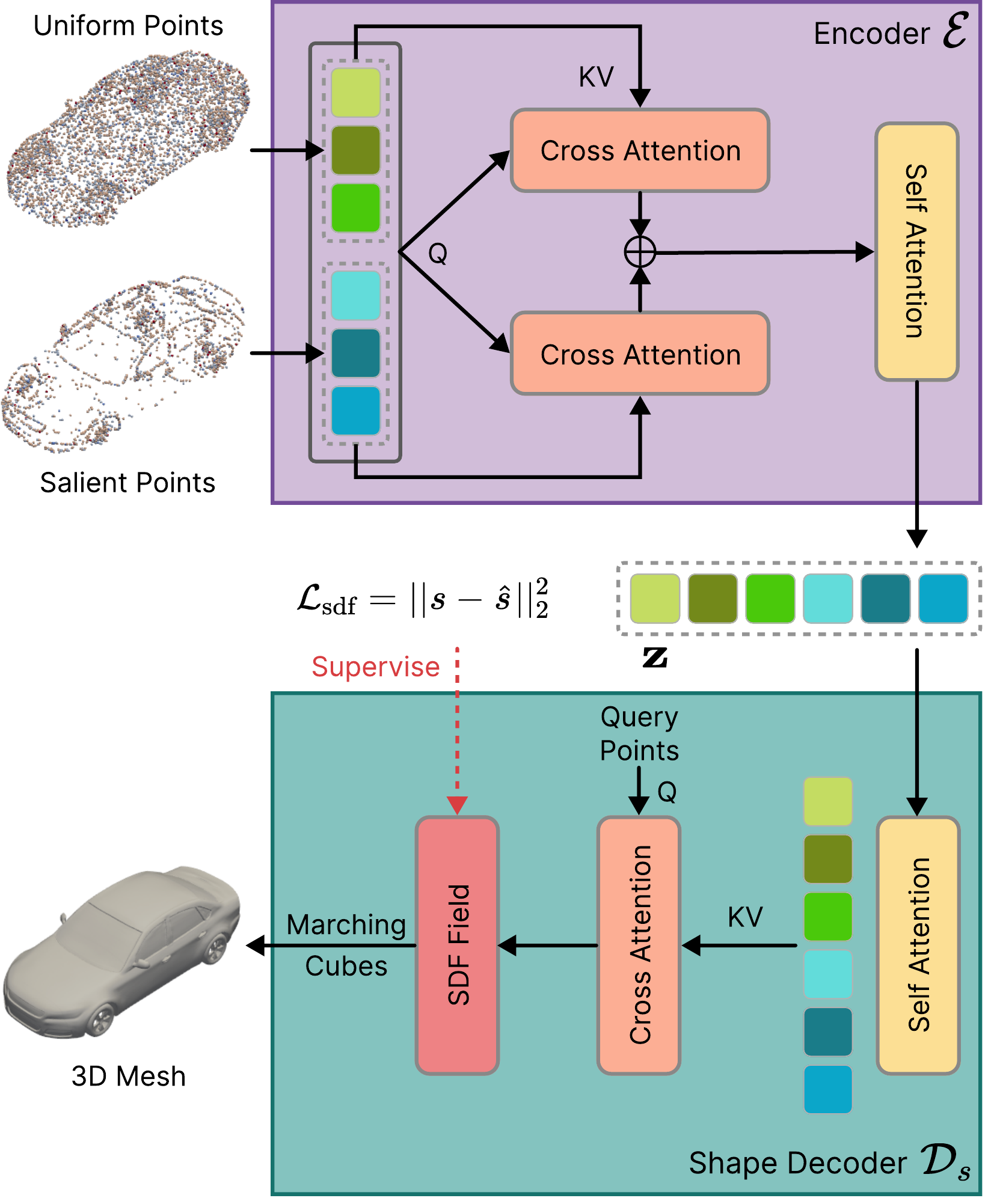}
    \caption{Overview of the SP-VAE shape encoder-decoder. The encoder fuses uniform and salient surface points via bidirectional cross-attention and self-attention to produce a latent code. The decoder predicts an SDF field from query points using cross-attention and reconstructs the mesh via marching cubes.}
    \label{fig:sup_vae}
\end{figure}
\subsection{Shape Encoder and Decoder}
We build our shape encoder-decoder architecture upon Dora~\cite{Chen25b}, while extending it to support SDF prediction~\cite{Li25c}, enabling finer geometric reconstruction than the original occupancy-based representation. As illustrated in Fig.~\ref{fig:sup_vae}, the mesh is first extracted into two complementary point sets:
(1) uniformly sampled surface points $\mathbf{P}_u$, which capture global coverage of the geometry, and
(2) salient edge points $\mathbf{P}_s$, which preserve high-curvature and structurally important regions.
These two sets provide separate geometric cues to the encoder. The encoder fuses them via bidirectional cross-attention, letting salient regions inform uniform samples and vice versa. The fused features then pass through self-attention layers to produce the latent code $\mathbf{z}$, capturing both coarse structure and fine details.
On the decoding side, we deviate from Dora’s occupancy-based formulation and instead predict an SDF field to better preserve high-frequency geometry. The latent code $\mathbf{z}$ is first enriched through several self-attention layers, and a set of 3D query points $\mathbf{x}\in\mathbb{R}^3$ is fed into a linear projection to form the attention queries. Through cross-attention between $\mathbf{x}$ and the latent features, the decoder estimates the corresponding signed distance value $s = \mathcal{D}_s(\mathbf{x},\mathbf{z})$, effectively conditioning the local geometry on the global shape embedding.
The predicted SDF field is then supervised with ground-truth distances sampled around the mesh, and the final mesh is extracted using the marching cubes algorithm~\cite{Lorensen98a}, yielding a high-quality reconstruction faithful to both global shape and local geometric details.

\begin{figure}[b]
    \centering
    \includegraphics[width=\linewidth]{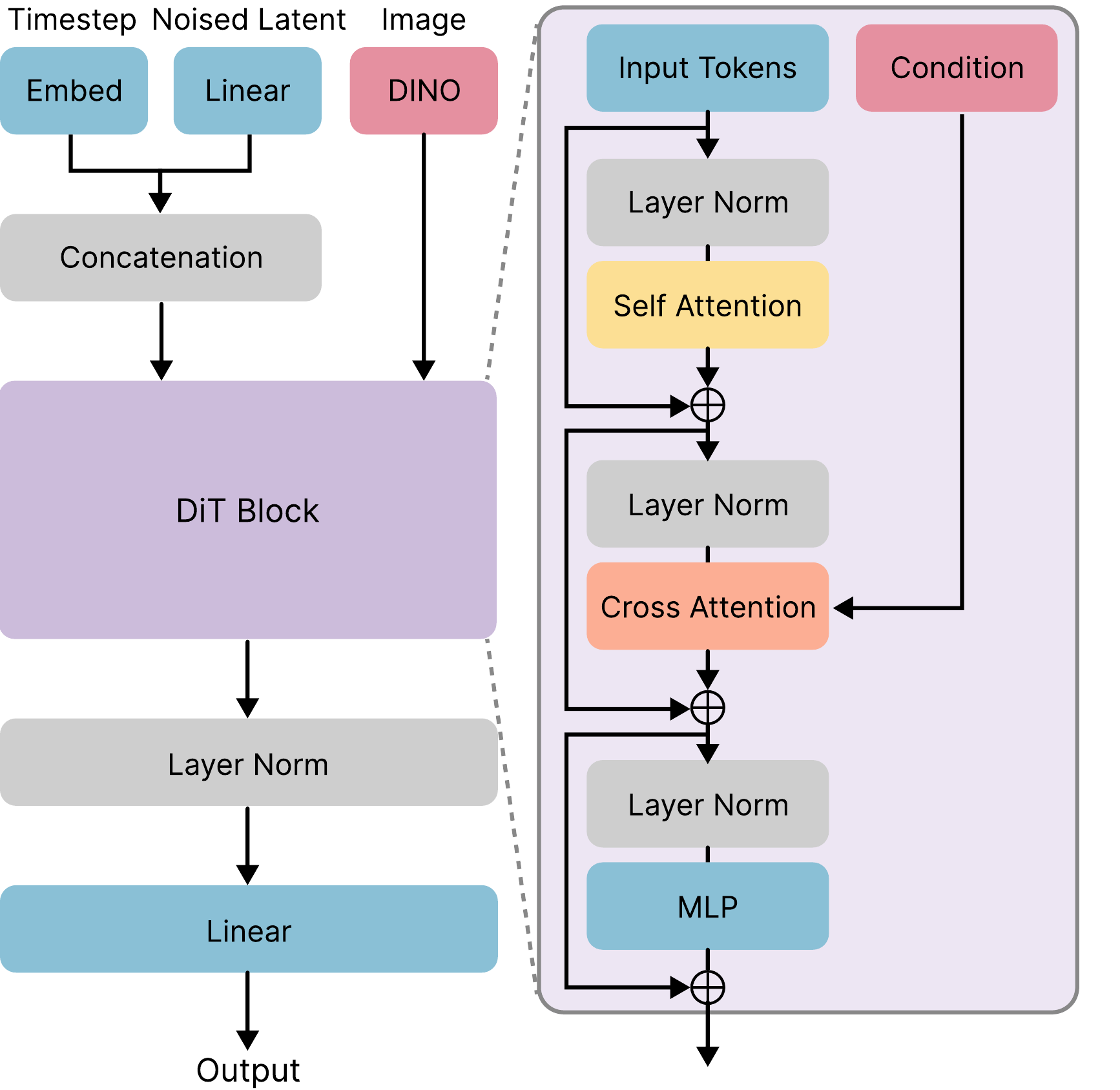}
    \caption{Diffusion Transformer (DiT) architecture. Noised latent and timestep embeddings form the input token sequence, while optional DINO-based conditioning is injected via cross-attention in each block. Each DiT block applies self-attention, cross-attention, and an MLP to produce the final velocity prediction.}
    \label{fig:sup_dit}
\end{figure}
\subsection{Diffusion Transformer Network}
We employ a Diffusion Transformer (DiT)~\cite{Peebles23a} within our flow-matching framework to parameterize the velocity field that transports noisy latent codes toward clean representations. As shown in Fig.~\ref{fig:sup_dit}, optional conditioning is introduced at the beginning of the network, where 
$\mathbf{c} = 
\left\{
  \begin{array}{ll}
    \textbf{I}, & \text{if conditional on image}, \\
    \varnothing, & \text{if unconditional}.
  \end{array}
\right.$ For image-conditioned generation, the input image is encoded by DINOv2~\cite{Oquab24a}, and the resulting feature tokens are embedded into a separate conditioning sequence that is injected into every DiT block via cross-attention. 
The noised latent $\mathbf{z}_{t_n}$ is mapped through a linear projection, while the timestep $t_n$ is encoded using a sinusoidal timestep embedder followed by an MLP. These two embeddings are concatenated to form the token sequence. Each DiT block adopts a pre-norm structure with residual connections and consists of self-attention over latent tokens, cross-attention with the conditioning tokens $\mathbf{c}$, and a feed-forward network. After all blocks, the final tokens are normalized and projected to produce the velocity field $\hat{\mathbf{u}}(\mathbf{z}_{t_n}, t_n, \mathbf{c})$ required by the flow-matching solver.

\section{Dataset details}
\noindent \textbf{DrivAerNet++}~\cite{Elrefaie24} is a large-scale aerodynamic design dataset comprising $8,000$ high-quality vehicle geometries, each accompanied by high-fidelity CFD simulations, including aerodynamic quantities such as drag coefficients, and both surface pressure and volumetric flow fields. It spans a wide range of automotive body styles, including fastback, notchback, and estateback, and features variations in underbody structure and wheel configurations. Our SP-VAE and flow-based generator are trained on this dataset.

\noindent \textbf{ShapeNet}~\cite{Chang15} car split is used to evaluate the generalization ability of our method, as it contains vehicle geometries that are not present in the training set. All shapes are uniformly rescaled to match the scale of DrivAerNet++. Although these meshes are physically imperfect yet geometrically reasonable, we use them as initial shapes for our \emph{Physics-Guided Shape Generation} pipeline. By optimizing from these initializations, our method refines the designs into physically efficient and aesthetically pleasing 3D shapes.


\section{Evaluation Metrics}
For {shape generation} quality, we evaluate geometric fidelity using F-score, Chamfer Distance, Accuracy, and IoU.

\subsection{Shape Generation}

\noindent \textbf{F-score.} {F-score ($\tau = 0.01$)} measures consistency between predicted mesh vertices $M$ and ground-truth vertices $G$, with threshold $\tau=0.01$:
\begin{equation}
\text{F-score}(\tau)
= 
2 \cdot 
\frac{
\text{Precision}(\tau)\cdot \text{Recall}(\tau)
}{
\text{Precision}(\tau)+\text{Recall}(\tau)
},
\end{equation}
where
\begin{equation}
\begin{aligned}
\text{Precision}(\tau)
&=
\frac{
|\{\, m\in M \mid d(m,G)<\tau \,\}|
}{
|P|
},
\\[4pt]
\text{Recall}(\tau)
&=
\frac{
|\{\, g\in G \mid d(g,M)<\tau \,\}|
}{
|G|
}.
\end{aligned}
\end{equation}
Here, $d(m,G)=\min_{g\in G}\|m-g\|_2$ denotes the nearest-neighbor distance from point $m$ to the ground-truth surface $G$.
Thus, $d(m,G)<\tau$ indicates that the point $m$ lies within a tolerance $\tau$ of the target surface.

\noindent \textbf{Chamfer Distance (CD).} CD measures the geometric discrepancy between the predicted point set $M$ and the ground-truth point set $G$. We use the bidirectional form, defined as:
\begin{equation}
\begin{aligned}
\text{CD}(M,G)
&=
\frac{1}{|M|}\sum_{m\in M}\min_{g\in G}\|m-g\|_2^2
\\
&+
\frac{1}{|G|}\sum_{g\in G}\min_{m\in M}\|g-m\|_2^2.
\end{aligned}
\end{equation}

\noindent \textbf{Accuracy (Coarse, Sharp, Overall).} Classification accuracy evaluates how well the predicted SDF-based inside/outside labels match the ground truth at sampled query points.
For a point $\mathbf{x}$, the predicted label is $y = \mathbf{1}[\, s(\mathbf{x}) \le 0 \,]$, where $s(\mathbf{x})$ is the predicted SDF. The ground-truth label is
$\hat{y} = \mathbf{1}[\, \hat{s}(\mathbf{x}) \le 0 \,]$, where $\hat{s}(\mathbf{x})$ is the ground truth SDF. For each sampling split $k \in \{\text{coarse}, \text{sharp}, \text{overall}\}$: 
\begin{equation}
    \text{Acc}_k
=
\frac{1}{N_k}
\sum_{i=1}^{N_k}
\mathbf{1}[y_i = \hat{y}_i].
\end{equation}
where coarse points are uniformly sampled within the bounding box, sharp points are generated by perturbing points around the ground-truth surface, and overall points are the union of the two.

\noindent \textbf{Intersection over Union (IoU).} Using the same binary inside–outside labels, IoU quantifies how well the predicted inside region overlaps with the ground-truth inside region. It is computed as the ratio between the number of points correctly classified as inside (intersection) and the number of points labeled as inside by either the prediction or the ground truth (union). A higher IoU indicates closer agreement between the predicted and true shape boundaries. For a given split $k \in \{\text{coarse}, \text{sharp}, \text{overall}\}$, IoU is computed as:
\begin{equation}
    \text{IoU}_k
=
\frac{
\sum_{i=1}^{N_k} \hat{y}_i \, y_i
}{
\sum_{i=1}^{N_k} \mathbf{1}[\hat{y}_i + y_i > 0] + \varepsilon
}.
\end{equation}

\subsection{Physical Estimation}
For the estimation task, we adopt standard regression metrics including Mean Squared Error (MSE), Mean Absolute Error (MAE), Maximum Absolute Error (Max AE), Relative $L_2$ Error (Rel L2), and Relative $L_1$ Error (Rel L1).
For each sampled point $i$, let $p_i$ denote the predicted pressure and $\hat{p}_i$ the ground-truth pressure, with $N$ being the total number of evaluated points.

\noindent \textbf{Mean Squared Error (MSE).} MSE measures the average squared deviation between predicted and true pressures, emphasizing larger errors more strongly:
\begin{equation}
\text{MSE} = 
\frac{1}{N}
\sum_{i=1}^{N}
(p_i - \hat{p}_i)^2.
\end{equation}

\noindent \textbf{Mean Absolute Error (MAE).} MAE computes the average absolute difference between predicted and ground-truth pressures, providing a more outlier-robust accuracy measure:
\begin{equation}
\text{MAE} = 
\frac{1}{N}
\sum_{i=1}^{N}
|p_i - \hat{p}_i|.
\end{equation}

\noindent \textbf{Maximum Absolute Error (Max AE).} Max AE Quantifies the worst-case prediction error by identifying the largest absolute pressure deviation across all points:
\begin{equation}
\text{Max AE} =
\max_{1 \le i \le N}
|p_i - \hat{p}_i|.
\end{equation}

\noindent \textbf{Relative $\mathbf{L_2}$ Error (Rel L2).} Rel L2 evaluates the global Euclidean discrepancy normalized by the magnitude of the ground-truth pressure field:
\begin{equation}
\text{Rel L2} =
\frac{
\lVert p - \hat{p} \rVert_2
}{
\lVert \hat{p} \rVert_2
},
\end{equation}
where $\lVert \cdot \rVert_2$ denotes the $\ell_2$ norm over all points.

\noindent \textbf{Relative $\mathbf{L_1}$ Error (Rel L1).} Rel L1 measures the normalized sum of absolute pressure errors relative to the total absolute ground-truth pressure:
\begin{equation}
\text{Rel L1}=
\frac{
\lVert p - \hat{p} \rVert_1
}{
\lVert \hat{p} \rVert_1
},
\end{equation}
where $\lVert \cdot \rVert_1$ is the $\ell_1$ norm, equal to the sum of absolute values over all points.

\section{CFD Simulation Setup in OpenFOAM}
\label{sec:sup_openfoam}
To evaluate the aerodynamic performance of our generated vehicle geometries, as shown in Table~\ref{tab:min_opt}, we perform high-fidelity CFD simulations using OpenFOAM~\cite{Jasak07} to compute the drag coefficient, surface pressure distributions, and airflow velocity fields. A uniform inlet freestream velocity of $30~\text{m/s}$, aligned with the vehicle’s longitudinal axis and directed toward the frontal surface, is prescribed to represent standard automotive aerodynamic operating conditions.
We employ the steady-state simpleFoam~\cite{Jasak07} solver together with the $k$-$\omega$ SST turbulence model. Each simulation is run on $32$ CPU cores for about $8$ hours and proceeds through $2500$ iterations to ensure convergence. The final aerodynamic drag coefficient is obtained by averaging the flow fields across the last $500$ iterations.


\begin{figure}[t]
    \centering
    \includegraphics[width=0.9\linewidth]{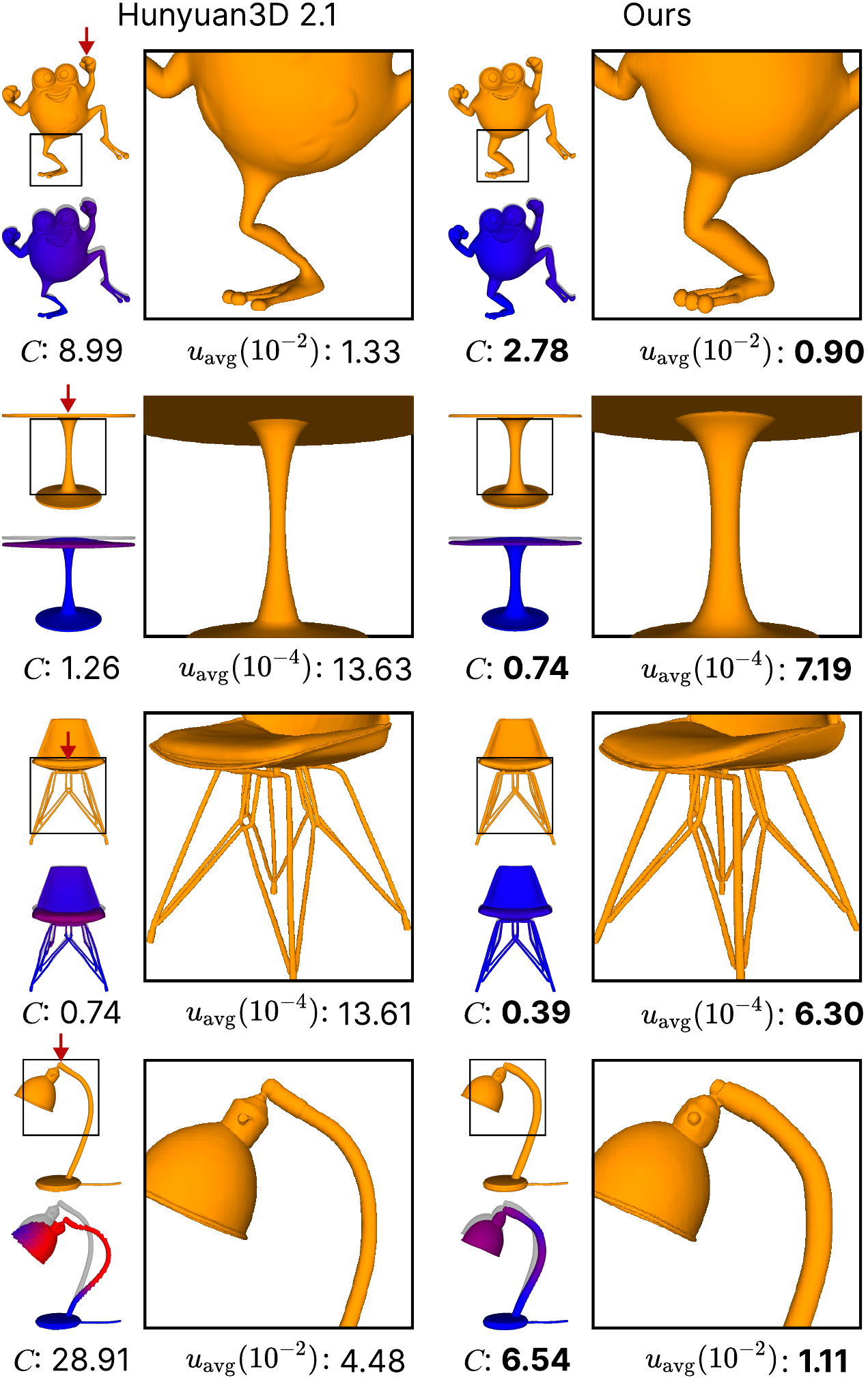}
    \caption{Additional results on structural optimization.}
    \label{fig:force_supp}
\end{figure}
\section{Generalization to Structural Optimization}

We further evaluate our physics-guided shape generation framework on \emph{structural optimization} by following the setting of PhysiOpt~\cite{Zhan25a}. Given external loads $\mathbf{f}$ and user-specified boundary conditions, we map the latent code $\mathbf{z}$ to an SDF representation, convert it into density-weighted finite elements, and then solve the linear static equilibrium equation:
\begin{equation}
\mathbf{K}(\mathbf{z})\,\mathbf{u} = \mathbf{f},
\end{equation}
where $\mathbf{K}$ is the stiffness matrix and $\mathbf{u}$ denotes the displacement field. As in PhysiOpt, the optimization objective is to reduce the compliance:
\begin{equation}
C = \mathbf{f}^{\top}\mathbf{u},
\end{equation}
which measures the structural deformation, or equivalently the strain energy, under the applied load. Lower compliance indicates a stiffer structure. We also report the average displacement:
\begin{equation}
u_{\text{avg}}=\frac{1}{N}\sum_{i=1}^{N}\|\mathbf{u}_i\|_2,
\end{equation}
where $\mathbf{u}_i$ is the displacement of the $i$-th FEM node. This formulation follows PhysiOpt’s differentiable FEM pipeline, which optimizes shapes directly in latent space under prescribed loads and boundary conditions. 

In this task, both PhysiOpt and our method are optimized for 180 steps. Specifically, we perform one velocity-based update after every 5 steps of physical refinement, so that shape plausibility and structural performance are improved jointly throughout the optimization. Fig.~\ref{fig:force_supp} presents additional structural optimization results of our method, where Hunyuan3D 2.1~\cite{hunyuan3d2025hunyuan3d} generates initial 3D shapes from images without physical awareness. It can be seen that our method improves physical performance while preserving shape quality, yielding structurally stronger and visually appealing shapes.

\end{document}